\documentclass[final,5p,times,twocolumn]{elsarticle}

\usepackage{amssymb}
\usepackage{bm}
\usepackage{amsmath}
\usepackage[linesnumbered,ruled]{algorithm2e}
\usepackage{enumitem} 
\usepackage{stfloats}
\usepackage{flushend}
\usepackage{subfig}
\usepackage{booktabs}
\usepackage{tabularx,threeparttable}
\usepackage[colorlinks,
linkcolor=black,
anchorcolor=black,
citecolor=blue,
urlcolor=black
]{hyperref}

\journal{Robotics and Autonomous Systems}

\begin{document}

\begin{frontmatter}

	\title{Distributed Motion Control of Multiple Mobile Manipulators for Reducing Interaction Wrench in Object Manipulation}
	
	\author[1]{Wenhang Liu}{}%
	\ead{liuwenhang@sjtu.edu.cn}
	
	\author[1]{Meng Ren}{}%
	\ead{meng\_ren@sjtu.edu.cn}
	
	\author[1]{Kun Song}{}%
	\ead{coldtea@sjtu.edu.cn}
	
	\author[1]{Gaoming Chen}{}%
	\ead{cgm1015@sjtu.edu.cn}
	
	\author[2,3]{Michael Yu Wang}{}%
	\ead{mywang@gbu.edu.cn}
	
	\author[1]{Zhenhua~Xiong\corref{cor1}}%
	\ead{mexiong@sjtu.edu.cn}

	\affiliation[1]{organization={School of Mechanical Engineering, State Key Laboratory of Mechanical System and Vibration, Shanghai Jiao Tong University, Shanghai, 200240}, country={China}}
	\affiliation[2]{organization={School of Engineering, Great Bay University, Guangdong, 523000}, country={China}}
	\affiliation[3]{organization={Department of Mechanical and Aerospace Engineering, The Hong Kong University of Science and Technology, Hong Kong, 999077}, country={China}}
	\cortext[cor1]{Corresponding author}

\begin{abstract}
In real-world cooperative manipulation of objects, multiple mobile manipulator systems may suffer from disturbances and asynchrony, leading to excessive interaction wrenches and potentially causing object damage or emergency stops.
Existing methods often rely on torque control and dynamic models, which are uncommon in many industrial robots and settings.
Additionally, dynamic models often neglect joint friction forces and are not accurate.
These methods are challenging to implement and validate in physical systems.
To address the problems, this paper presents a novel distributed motion control approach aimed at reducing these unnecessary interaction wrenches.
The control law is only based on local information and joint velocity control to enhance practical applicability.
The communication delays within the distributed architecture are considered.
The stability of the control law is rigorously proven by the Lyapunov theorem.
In the simulations, the effectiveness is shown, and the impact of communication graph connectivity and communication delays has been studied.
A comparison with other methods shows the advantages of the proposed control law in terms of convergence speed and robustness.
Finally, the control law has been validated in physical experiments.
It does not require dynamic modeling or torque control, and thus is more user-friendly for physical robots.
\end{abstract}
	
	
\end{frontmatter}

%% \linenumbers

%% main text
\section{Introduction}
Mobile manipulator robots combine the extensive mobility of a mobile platform with the dexterous manipulation capabilities of a robotic manipulator.
This integration allows them to be used in various industrial applications, including object delivery, assembly, machining, and transportation \cite{thakar2020manipulator,jang2023motion,tao2019mobile}.
As the complexity of industrial tasks rises, the limitations of single robots become apparent.
Tasks such as manipulating large objects and performing complex assemblies often require the collaboration of multiple robots \cite{dogar2015multi,tao2022robotic}.
Moreover, multi-robot collaboration offers more advantages, such as efficiency, robustness, adaptivity, and flexibility \cite{feng2020overview}.
Therefore, in industrial automation, Multiple Mobile Manipulator Systems (MMMS) have attracted widespread attention in object manipulation.

In related research, some studies focus on motion planning problems, mainly including redundant and coordinated planning \cite{cui2024task}.
However, in practice, it is impossible to perfectly achieve the planned results in robot motions due to various real-world factors.
These factors include disturbances and asynchrony between the robots, which may result from communication delays or kinematic uncertainties.
It can lead to deviations from the planned cooperative performance.
Thus, some research aims at cooperative control of MMMS to ensure the desired performance \cite{sugar2002control}.
To track the reference trajectory and avoid obstacles, a method integrating formation control with constrained optimization for MMMS was proposed in \cite{wu2021distributed}.
A constrained optimization method for multi-robot formation control in dynamic environments was proposed in \cite{alonso2017multi}, optimizing formation parameters to avoid collisions and progress toward goals.
The approach included local motion planning via sequential convex programming and global path planning through sampling and graph search.
The above studies enable MMMS to achieve some desired coordination performance.

In object manipulation, the end-effector of the robot is usually a gripper and grasps the object.
When multiple robots grasp an object simultaneously, it necessitates precise coordination of interaction wrenches between robots and the object to ensure effective and safe collaboration.
Due to the tight coupling, any small imbalance or misalignment may cause excessive interaction wrenches by mutual pushing and pulling.
The increased wrenches may damage the object or cause the object to slip \cite{sieber2018human,marino2017distributed}.
Therefore, in object manipulation, the coordination of interaction wrenches is a crucial aspect of collaborative performance.

Generally, the control strategies of MMMS in maintaining desired interaction wrenches can be categorized into centralized, leader-follower, and fully distributed methods.
Centralized methods employ a central controller to control the movements and motions of all mobile manipulators \cite{li2008robust}.
These methods benefit from a global perspective, enabling optimal task allocation and coordination strategies.
However, centralized methods suffer from scalability issues as the number of robots increases \cite{andaluz2011coordinated}.
The computational burden on the central controller grows, and the system becomes vulnerable to a single point of failure.
In recent years, more attention has been given to the latter two methods in research.

The leader-follower method simplifies coordination by assigning one robot as a leader and the rest as followers.
Leaders make key decisions, while followers adjust their motions based on the leaders' states \cite{wang2016force,chen2018cooperative}.
In \cite{bechlioulis2018collaborative}, a leader-follower structure was used for MMMS to cooperatively transport objects in a constrained workspace.
In \cite{ren2024hybrid}, the coordination was ensured by maintaining the relative pose relationship of the fixed end-effectors between the coupled mobile manipulators.
In the simulation or experiment, a robot was chosen as the leader.
Similarly, leader-follower control methods can also be found in \cite{liu2024cooperative,wu2016collaboration,kume2007coordinated}.
While the leader-follower structure reduces control complexity for followers and is practical for real-world implementation, it has notable drawbacks.
The system's performance depends heavily on the leader's reliability and decision-making, making it vulnerable to single points of failure similar to centralized methods.

Fully distributed methods represent the most advanced approach, where each mobile manipulator makes decisions based on local information and communication with neighboring robots.
In \cite{ren2020fully}, a distributed adaptive controller was proposed to achieve synchronization based on the desired load distribution and desired internal force.
The effectiveness of their method was validated through experiments.
A distributed model predictive control method was presented for MMMS to transport objects with respect to state, control input, and synchronization constraints in \cite{qin2022cooperation}.
The method simplifies the problem into two independent synchronization tracking control issues for task-space end-effectors and null-space mobile bases.
In \cite{ponce2016cooperative}, a chatterless integral sliding mode force-position control was proposed.
Each robot was designed to correct collectively the velocity deviation due to the pushing of the other robots.
Distributed coordination and cooperation control for networked mobile manipulators over a topology was introduced in \cite{dai2016distributed}.
Synchronization control combined with force controls ensured cooperative manipulation and transportation.
In \cite{verginis2019robust}, a control approach was proposed to guarantee predefined transient and steady-state performance for object trajectory.
Load distribution between robots was also considered by using a grasp matrix pseudoinverse.

\begin{table*}[!t]
	\caption{Comparison between distributed control methods.}
	\label{tab:summary}
	\centering
	\begin{tabular}{m{0.15\textwidth}<{\centering}m{0.15\textwidth}<{\centering}m{0.15\textwidth}<{\centering}m{0.2\textwidth}<{\centering}} \toprule
		& Torque Control & Verification & Positioning  System \\ \midrule
		\cite{qin2022cooperation,ponce2016cooperative,yan2021decentralized,dai2016distributed,verginis2019robust}   & Need & Simulation & -  \\
		\cite{wu2021distributed} & No need & Simulation & - \\
		\cite{ren2020fully,xu2023reinforcement} & Need & Experiment & Need \\
		ours & No need & Experiment & No need \\ \bottomrule
	\end{tabular}
\end{table*}

The research on distributed control is summarized in Table \ref{tab:summary}.
It can be seen that most methods require torque control.
These methods establish dynamic models to estimate the output wrenches based on the Newton-Euler method or the Lagrangian method \cite{zhang2023hybrid}.
However, in complex multi-degree-of-freedom systems, determining the relevant dynamic parameters presents significant challenges.
In these systems, parameters such as mass, inertia, and joint stiffness are critical for control, but they are often difficult to measure or estimate with high precision.
Small errors in these parameters can lead to large discrepancies and unwanted performance.
Furthermore, joint friction is often ignored or oversimplified in dynamic models.
These factors make it difficult to validate their methods in physical systems \cite{qin2022cooperation,ponce2016cooperative,yan2021decentralized,dai2016distributed,verginis2019robust}.
In a few research with experimental tests, external positioning systems were used to obtain the motion information of the end-effectors \cite{ren2020fully,xu2023reinforcement}.

When an accurate dynamic model is unavailable and torque control is not feasible, the main challenge is how to develop an effective control law with less information.
Besides, the design of this control law should be as simple as possible, ensuring that it can be applied in most conditions.
Therefore, to address these shortcomings and make MMMS more convenient and applicable in real-world object manipulation, this paper presents a distributed motion control strategy.
The robot obtains states through a local sensor and information from neighboring robots.
Then, the control inputs for joint velocities are computed to reduce interaction wrenches.
In the control process, neither an external positing system nor torque control is required.
The stability of the control law has been theoretically proven through the Lyapunov theorem.
In the simulations, the effects of graph connectivity and time delays on the control law have been studied.
A comparison with other methods has also been made, demonstrating the advantages of the proposed control law.
Finally, the control law is validated through physical experiments.

The main contributions of this paper are twofold.
First, the proposed control law requires very few conditions, only local force-torque information. Compared to most methods based on dynamic modeling, it does not need torque control, which offers better versatility for different robots, especially for industrial robots without accurate dynamic models.
Second, the proposed control law has been validated in physical systems and proven to be robust to communication delays. Most existing studies only provide theoretical proofs and simulations, with no physical validation.

The remainder of this paper is organized as follows.
In Section II, some basic concepts are defined.
In Section III, the problem is formulated and the proposed control law is presented. Also, the method is proven theoretically and verified by simulations. 
In Section IV, experiments are conducted on physical robots and results are analyzed.
Finally, Section V concludes the paper and outlines future work.

\section{Preliminaries}

\subsection{Kinematics of Mobile Manipulator}

\begin{figure}[t]
	\includegraphics[width=\columnwidth]{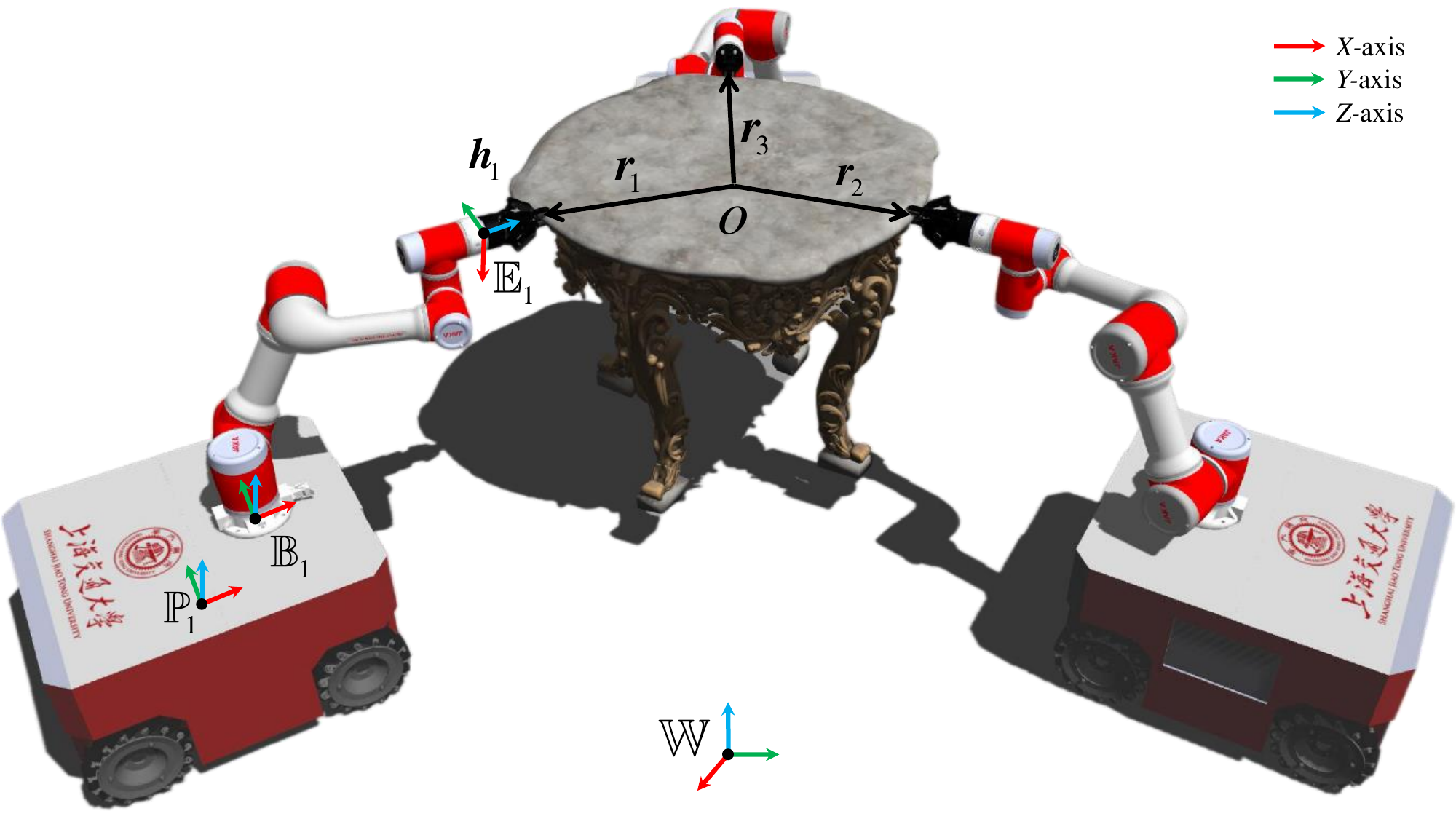}
	\centering
	\caption{
		Three mobile manipulator robots collaboratively manipulate an object. Some basic coordinate frames are labeled.
	}
	\label{MMMS}       
\end{figure}

A mobile manipulator robot usually consists of a mobile platform and a mounted robotic manipulator.
This paper considers an omnidirectional mobile platform and a six-degree-of-freedom manipulator.
As shown in Fig.~\ref{MMMS}, three mobile manipulators cooperatively manipulate objects.
$\mathbb{W}$ denotes the frame of the world.
$\mathbb{P}_i$, $\mathbb{B}_i$, and $\mathbb{E}_i$ denote the frames of the mobile platform, the base of the manipulator, and the end-effector of the $i$-th mobile manipulator, respectively, $i = 1, \cdots, n$.
Let $\bf{T}$, $\bf{R}$, $\bm{p}$ represent the relationship between different frames.
For example, ${^{\mathbb{W}}_{\mathbb{E}_i}}{\bf{T}} \in SE(3)$ represents the pose of frame $\mathbb{E}_i$ in frame $\mathbb{W}$, ${^{\mathbb{W}}_{\mathbb{E}_i}}{\bf{R}} \in SO(3)$ represents the orientation of frame $\mathbb{E}_i$ in frame $\mathbb{W}$, and ${^{\mathbb{W}}_{\mathbb{E}_i}}{\bm{p}} \in \mathbb{R}^3$ represents the translation of frame $\mathbb{E}_i$ in frame $\mathbb{W}$.
Based on frame transformations, the following fundamental relationships can be obtained
\begin{equation}
	\label{transformations}
	^{\mathbb{W}}_{\mathbb{E}_i}{\bf{T}} = {^{\mathbb{W}}_{\mathbb{P}_i}}{\bf{T}} \ {^{\mathbb{P}_i}_{\mathbb{B}_i}}{\bf{T}} \ {^{\mathbb{B}_i}_{\mathbb{E}_i}}{\bf{T}}
\end{equation}
It should be noted that ${^{\mathbb{P}_i}_{\mathbb{B}_i}{\bf{T}}}$ is invariant and solely depends on the location of the manipulator mounted on the mobile platform.

The forward kinematics of the robot is 
\begin{equation}
	\label{kinematics}
	\bm{e}_i = f_{\text{mm}}(\bm{q}_{\text{mm},i})
\end{equation}
where $\bm{e}_i \in \mathbb{R}^6$ is the pose of the end-effector in frame $\mathbb{W}$, $f_{\text{mm}}(\cdot)$ is related to ${^{\mathbb{W}}_{\mathbb{E}_i}}{\bf{T}}$, and $\bm{q}_{\text{mm},i} \in \mathbb{R}^9$ is the configuration of the robot.
$\bm{q}_{\text{mm},i} = [\bm{q}_{\text{p},i}^{\text{T}}, \bm{q}_{i}^{\text{T}}]^{\text{T}}$, specifically, $\bm{q}_{\text{p},i} = [x, y, \alpha]^{\text{T}} \in \mathbb{R}^3$ determining ${^{\mathbb{W}}_{\mathbb{P}_i}}{\bf{T}}$ denotes the position and orientation of the mobile platform, and $\bm{q}_{i} = [\theta_1, \theta_2, \theta_3, \theta_4, \theta_5, \theta_6]^{\text{T}} \in \mathbb{R}^6$ determining ${^{\mathbb{B}_i}_{\mathbb{E}_i}}{\bf{T}}$ denotes the joint angles of the manipulator.
Nonholonomic constraints are not considered in this paper.
Thus, differentiating Eq.(\ref{kinematics}), the velocity relationship is
\begin{equation}
	\label{Jacobian}
	\dot{\bm{e}}_i = \bm{J}_{\text{mm},i}(\bm{q}_{\text{mm},i})\dot{\bm{q}}_{\text{mm},i} = [\bm{J}_{\text{p},i}, \bm{J}_{i}][\bm{q}_{\text{p},i}^{\text{T}}, \bm{q}_{i}^{\text{T}}]^{\text{T}}
\end{equation}
where $\bm{J}_{\text{mm},i} = [\bm{J}_{\text{p},i}, \bm{J}_{i}]$ is the Jacobian matrix of the $i$-th robot.
$\bm{J}_{\text{p},i}$ and $\bm{J}_{i}$ are the Jacobian matrices of the $i$-th mobile platform and manipulator, respectively.

\subsection{Force-Torque Sensor}

The manipulator is equipped with a gripper and a force-torque sensor.
The data measured by force-torque sensors are usually in their own coordinate systems.
Let $^{\mathbb{E}_i}\bm{h}_{i} = [{^{\mathbb{E}_i}}\bm{f}_{i}^{\text{T}},{^{\mathbb{E}_i}}\bm{t}_{i}^{\text{T}}]^{\text{T}}$ represent the data measured by the sensor at the end effector of the $i$-th robot.
$\bm{h}_{i} = [\bm{f}_{i}^{\text{T}},\bm{t}_{i}^{\text{T}}]^{\text{T}}$ represents the force and torque exerted on the end-effector of the $i$-th robot in frame $\mathbb{W}$.
According to the force transfer relationship, it can be derived
\begin{equation}
	\label{force_snsor}
	\begin{aligned} 
		\bm{f}_{i} = {^{\mathbb{W}}_{\mathbb{E}_i}}{\bf{R}} \cdot {^{\mathbb{E}_i}}\bm{f}_{i} \quad
		\bm{t}_{i} = {^{\mathbb{W}}_{\mathbb{E}_i}}{\bf{R}} \cdot {^{\mathbb{E}_i}}\bm{t}_{i}
	\end{aligned}
\end{equation}

Before measuring data, it is necessary to perform calibration and gravity compensation for the sensor.
In unloaded conditions, the force data consists of
\begin{equation}
	\label{force_calibration}
	\begin{aligned} 
		{^{\mathbb{E}_i}}\bm{h}_{i} = \left[ \begin{matrix} {^{\mathbb{E}_i}}\bm{f}_{i}  \\	 {^{\mathbb{E}_i}}\bm{t}_{i}  \\	\end{matrix}\right] = \left[ \begin{matrix} \bm{f}_{\text{s}\_0}  \\	 \bm{t}_{\text{s}\_0}  \\	\end{matrix}\right] + \left[ \begin{matrix} \bm{f}_{\text{s}\_g}  \\	 \bm{t}_{\text{s}\_g}  \\	\end{matrix}\right]
	\end{aligned}
\end{equation}
where $\bm{f}_{\text{s}\_0}$ and $\bm{t}_{\text{s}\_0}$ are the drift errors, $\bm{f}_{\text{s}\_g}$ and $\bm{t}_{\text{s}\_g}$ are caused by the gravity of the gripper $mg$.
The displacement from the sensor to the gripper's center of mass is denoted as ${^{\mathbb{E}}_{g}}{\bm{p}}$.
Parameters $\bm{f}_{\text{s}\_0}$, $\bm{t}_{\text{s}\_0}$,  ${^{\mathbb{E}}_{g}}{\bm{p}}$, and $mg$, which need further calibration, are linearly independent.
Therefore, these parameters can be calculated using the least squares method after obtaining enough sensor data in unloaded conditions \cite{zhou2023robotic}.

The total wrench applied to the object by robots is
\begin{equation}
	\label{total_wrench}
	\begin{aligned} 
		\bm{h}_{o} &= -\bm{G} \left[ \begin{matrix} \bm{h}_{1}^{\text{T}}  &	\cdots & \bm{h}_{n}^{\text{T}}	\end{matrix}\right]^{\text{T}} \\
		\bm{G} & = \left[ \begin{matrix} I_{3} & 0_{3} & \cdots & I_{3} & 0_{3}\\	 [\bm{r}_1]_{\times} &  I_{3} & \cdots &   [\bm{r}_n]_{\times} &  I_{3}  \end{matrix}\right]
	\end{aligned}
\end{equation}
where $\bm{G}_{6\times6n}$ is the grasp matrix, which is only related to the relative poses of the grasping points and the object's center of mass \cite{erhart2016model,erhart2015internal, ren2020fully}, as the $\bm{r}_1, \bm{r}_2, \bm{r}_3$ shown in Fig.~\ref{MMMS}.
$[\bm{r}_i]_{\times}$ represents the skew-symmetric matrix of $\bm{r}_i$.

\subsection{Graph Theory}

The communication structure of a system with $n$ robots can be typically represented on graphs.
Let $\mathcal{G=(V,E)}$ denote a graph, where $\mathcal{V} = \left\{ v_{i} \right\}$ is the vertex set and $\mathcal{E} = \left\{ a_{ij} \right\}$ is the edge set.
$a_{ij} = 1$ if the $i$-th robot can receive information from the $j$-th robot, otherwise $a_{ij} = 0$.
The adjacency matrix of the graph is $A = (a_{ij}) \in \mathbb{R}^{n\times n}$.
The degree matrix of the graph is $D = \text{diag}\left\{   \sum_{j=1}^{n} a_{ij}   \right\} \in \mathbb{R}^{n\times n}$.
The Laplacian matrix of the graph is $L = D - A$.
In this paper, the directed graph is considered, which denotes $a_{ij} \neq a_{ij}$.

\section{The Proposed Control Method}

\subsection{Problem Statement and Control Law}

\textit{Assumption 1}: The robot grasps the object tightly with its gripper, ensuring no slippage and sufficient force transmission during motion.

\noindent \textit{Assumption 2}: In the control process, the manipulator avoids encountering singular configurations.

\noindent \textit{Assumption 3}: Each robot receives information from at least one other robot, which is $\sum_{j=1}^{n} a_{ij}  \geq 1 $. The upper bound for communication delay between robots is $\tau$.

\textit{Assumption 1} is a very general assumption, which can also be found in \cite{ren2020fully,zhang2023hybrid,xu2023distributed,verginis2022cooperative,carey2021collective}.
\textit{Assumption 2} can be also found in \cite{li2008robust,ren2024hybrid,dai2016distributed}.
\textit{Assumption 3} is related to the stability proof and will be further discussed in subsequent theoretical proofs.

This paper considers the addition of a compensatory motion to the original motions to achieve the desired interaction wrenches.
For the sake of brevity, $\bm{q}(t)$ is abbreviated as $\bm{q}$ in the absence of time delay.
For each robot, assuming that $\bm{h}_{i}^{\text{ref}}$ and $\bm{q}_{\text{mm},i}^{\text{ref}}$ have been planned.
In planning, the desired interaction wrenches are achieved if $\dot{\bm{q}}_{\text{mm},i}^{\text{ref}}$ is the input.
However, due to model deviations and disturbances, this ideal situation is almost impossible.
If global positioning and velocity information could be obtained, the control process would be relatively easy since the actual velocity is corrected to match the desired velocity.
However, this often requires high-precision and large-scale positioning systems, which are often difficult to implement \cite{xu2023reinforcement}.
Therefore, information from local sensors is considered to design the control laws.
Besides, it is also important to consider the communication delay between robots as it is a very common situation in reality.
Let $\tau_{ij}(t)$ denote the time delay in the $i$-th robot receiving information from the $j$-th robot.
In this paper, the most general delays are considered.
The communication delays are non-uniform, asymmetric, time-varying, and unknown.
For example, $\tau_{12}(t) \neq \tau_{21}(t)$, $\tau_{12}(t) \neq \tau_{13}(t)$, and thus robots cannot achieve synchronization by querying their own historical information when receiving unknown communication delays.

The actual force exerted on the $i$-th robot during motion by the object is
\begin{equation}
	\label{stiffness}
	\begin{aligned} 
		\bm{h}_i - \bm{h}_i^{\text{ref}} &= K(\sum_{j=1, j\neq i}^{n}   (\Delta\bm{e}_{j}  -    \Delta\bm{e}_{i} ) )  \\
		\quad \Delta\bm{e}_{i} &= \bm{e}_{i} - \bm{e}_{i}^{\text{ref}}
	\end{aligned}
\end{equation}
where $K_{6\times 6} > 0$ is a diagonal matrix, representing the stiffness of the object or external environment \cite{sugar2002control,chen2024compliance,zhang2023hybrid}.
Eq.(\ref{stiffness}) can be rewritten as
\begin{equation}
	\label{correction}
	\bm{h}_i - \bm{h}_i^{\text{ref}} = K( \bm{e}_i^{\text{d}} -  \bm{e}_i   ) 
\end{equation}
where $\bm{e}_i^{\text{d}}$ is the desired correction target.
Through the force-torque sensor, the direction of correction can be obtained.

\begin{figure}[t]
	\includegraphics[width=\columnwidth]{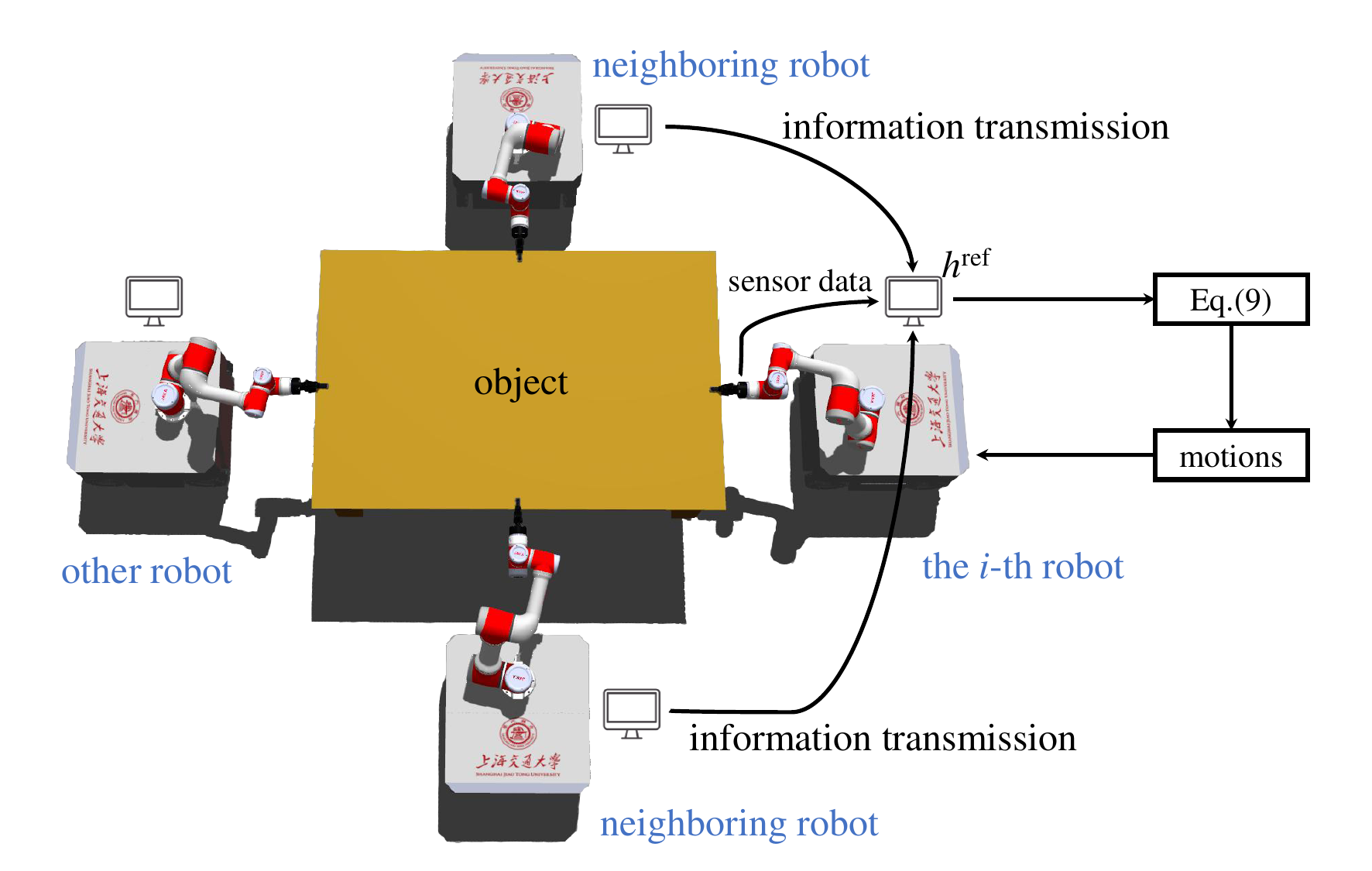}
	\centering
	\caption{
		The overall framework. The control process of the $i$-th robot is illustrated.
	}
	\label{framework}       
\end{figure}

In the control of mobile manipulators, redundancy may be introduced.
It can be considered when there are additional tasks, such as obstacle avoidance \cite{wu2021distributed,chen2018cooperative,ren2024hybrid}, which is not the focus of this paper.
Since the range of control correction is generally small and \textit{Assumption 2} holds, therefore, only the motions of manipulators are used for correction in the control process \cite{liu2024cooperative,wu2016collaboration,zhang2023hybrid}.
Moreover, the accuracy of the manipulator is higher than the platform, making it worth prioritizing the motion of the manipulator \cite{yamazaki2022approaching}.
Considering $\dot{\bm{q}}_i = \bm{u}_i$ and the control law $\bm{u}_i$ is designed as follows
\begin{equation}
	\label{control_law}
	\begin{aligned} 
		\bm{u}_{i}  =   - k & \sum_{j = 1}^{n}a_{ij} \Big[ f^{-1}(K^{-1}( \bm{h}_i^{\text{ref}} - \bm{h}_i ))  \\
		& -  \beta f^{-1}(K^{-1}( \bm{h}_j^{\text{ref}}(t - \tau_{ij}) - \bm{h}_j(t - \tau_{ij}) )) \Big] 
	\end{aligned}
\end{equation}
where $k$ and $\beta$ are positive coefficients, and $f^{-1}(\cdot)$ is the inverse kinematics of the manipulator.
Eq.(\ref{control_law}) computes the joint motion of the manipulator to reduce interaction wrenches.
The information required for this control law is the local sensor information, the current joint position of the manipulator, and the information from neighboring robots.
The control law ensures that the end-effector of the robot moves towards the desired target $\bm{e}_i \to \bm{e}_i^{\text{d}}$, thereby maintaining the interaction wrench at the expected value $\bm{h}_i \to \bm{h}_i^{\text{ref}}$.
The control framework is shown in Fig.~\ref{framework}, which details how one robot performs feedback control based on the required information.
Since each robot individually computes its own motions, without the need for a central controller to compute the motions for all robots, the control law is distributed.

Eq.(\ref{correction}) can be rewritten in the joint space as 
\begin{equation}
	\label{joint_space}
	\bm{h}_i - \bm{h}_i^{\text{ref}}   =          K( f(\bm{q}_i^{\text{d}}) - f(\bm{q}_i))
\end{equation}
where $\bm{q}_i^{\text{d}}$ is the desired correction joint target.
Substituting Eq.(\ref{joint_space}) into Eq.(\ref{control_law}), the control law can be rewritten as
\begin{equation}
	\label{joint_law}
	\begin{aligned} 
		\bm{u}_{i} =  - k \sum_{j = 1}^{n}{a_{ij} ((\bm{q}_i - \bm{q}_i^{\text{d}}) - \beta(\bm{q}_j(t - \tau_{ij}) - \bm{q}_j^{\text{d}}(t - \tau_{ij})) )   }        \\
	\end{aligned}
\end{equation}
Eq.(\ref{control_law}) is the control law used in actual calculations, indicating the input and output. 
The control law is written as Eq.(\ref{joint_law}) to conveniently illustrate subsequent proofs of stability.

\subsection{Stability Analysis}

\textit{Lemma 1} (Jensen Inequality):  For a real convex function $\varphi(\cdot)$, $x_i, i \in [1,n]$ are in its domain, it can be derived that
\begin{equation}
	\label{Jensen}
	\varphi(\frac{\sum_{i=1}^{n}x_i}{n})  \leq \frac{\sum_{i=1}^{n}\varphi(x_i)}{n}
\end{equation}

\noindent \textit{Lemma 2} (Quadratic Integral Inequality):  For a symmetric positive definite matrix $R_{n\times n}$ and a differentiable $n$-dimension vector function $\bm{x}(t): [a,b] \to \mathbb{R}^n$, it can be derived that
\begin{equation}
	\label{Integral}
	\int_{a}^{b}{ \dot{\bm{x}}(s)^{\text{T}}R\dot{\bm{x}}(s) \text{d}s } \geq \frac{(\bm{x}(b)-\bm{x}(a))^\text{T}R(\bm{x}(b)-\bm{x}(a))}{b-a}
\end{equation}

\noindent \textit{Lemma 3} \cite{park2011reciprocally}:  Let $\varphi_1, \varphi_2, ..., \varphi_n: \mathbb{R}^{m} \to \mathbb{R}$ have positive values in an open subset $S \in \mathbb{R}^{m}$, it can be derived that
\begin{equation}
	\label{lower_bound}
	\begin{aligned} 
		\mathop{\min}_{\left\{\alpha_i|\alpha_i>0, \sum_{i}^{}\alpha_i=1\right\}} \sum_{i}^{}\frac{1}{\alpha_i}\varphi_{i}(t) = \sum_{i}^{}\varphi_{i}(t) + 	\mathop{\max}_{g_{i,j}(t)}\sum_{i\neq j}g_{i,j}(t) \\
		\text{s.t.}  \begin{cases}  \quad g_{i,j}(t): \mathbb{R}^{m} \to \mathbb{R}, \ g_{i,j}(t) = g_{j,i}(t) \\\\ \left[\begin{matrix} \varphi_{i}(t) & g_{i,j}(t)  \\	g_{i,j}(t) & \varphi_{j}(t)  \\	\end{matrix}\right] \geq 0	\end{cases}			\end{aligned}
\end{equation}

Some fundamental relationships are as follows.
\begin{equation}
	\label{sum}
	\sum_{j=1}^{n}a_{ij} \leq n , \sum_{j=1}^{n}a_{ji} \leq n, \ \sum_{i=1}^{n}\sum_{j=1}^{n}a_{ji}b_{j} = \sum_{i=1}^{n}\sum_{j=1}^{n}a_{ij}b_{i} 
\end{equation}

The joint position errors of the $i$-th robot are 
\begin{equation}
	\label{error}
	\bm{\xi}_{i} = \bm{q}_i - \bm{q}_i^{\text{d}}
\end{equation}
Taking the time derivative of Eq.(\ref{error}) and substituting the control law Eq.(\ref{joint_law}) into $\dot{\bm{q}}_i$ obtains
\begin{equation}
	\label{error_dot}
	\dot{\bm{\xi}}_{i} = -k\sum_{j = 1}^{n}{a_{ij}}{\bm{\xi}}_{i} + k\beta\sum_{j = 1}^{n}{a_{ij} \bm{\xi}_{j}(t-\tau_{ij})   }
\end{equation}
For the sake of brevity, the quadratic type $\bm{\xi}_{i}^{\text{T}}\bm{\xi}_{i}$ is abbreviated as $\bm{\xi}_{i}^2$.
Considering the following Lyapunov function
\begin{equation}
	\label{Lyapunov}
	\begin{aligned} 
		V(t) &= V_1 +  V_2 + V_3       \\
		V_1 &= \gamma_1\sum_{i=1}^{n} \bm{\xi}_{i}^2 \\
		V_2 &= \gamma_2\sum_{i=1}^{n}\sum_{j = 1}^{n}{ a_{ji}  \int_{t-\tau}^{t}{ \bm{\xi}^2_{i}(s)}}\text{d}s  \\
		V_3 &= \gamma_3\sum_{i=1}^{n}\sum_{j = 1}^{n}{ a_{ji} \int_{-\tau}^{0}\int_{t-\omega}^{t}{ \dot{\bm{\xi}}^2_{i}(s)}}\text{d}s\text{d}\omega
	\end{aligned}
\end{equation}
where $\gamma_1, \gamma_2$ and $\gamma_3$ are the positive coefficients of the Lyapunov function.
The Lyapunov function is an equation based on the joint position errors and is positive definite.
Taking the time derivative of the Lyapunov function obtains
\begin{equation}
	\label{Lyapunov_dot}
	\begin{aligned} 
		\dot{V}_1 &= \gamma_1 \sum_{i=1}^{n} 2 \bm{\xi}_{i}^{\text{T}}\dot{\bm{\xi}}_{i}  \\
		\dot{V}_2 &= \gamma_2 \sum_{i=1}^{n}\sum_{j = 1}^{n}{ a_{ji}  { (\bm{\xi}_{i}^2 - \bm{\xi}_{i}^2(t-\tau)) }}  \\
		\dot{V}_3 &= \gamma_3 \sum_{i=1}^{n}\sum_{j = 1}^{n} a_{ji}  {      (\tau \dot{\bm{\xi}}_{i}^2 -     \int_{t-\tau}^{t}{ \dot{\bm{\xi}}_{i}^2(s) }  \text{d}s    )  }
	\end{aligned}
\end{equation}

Substituting Eq.(\ref{error_dot}) into $\dot{V}_1$ obtains
\begin{equation}
	\label{Lyapunov_1}
	%	\begin{aligned} 
		%	\dot{V}_1 &= -2\gamma_1 k \sum_{i=1}^{n}\sum_{j = 1}^{n}{a_{ij}}  \bm{\xi}_{i}^2 + 2\gamma_1k\beta \sum_{i=1}^{n}\sum_{j = 1}^{n}{a_{ij}}  \bm{\xi}_{i}^{\text{T}} \bm{\xi}_{j}(t-\tau_{ij}) \\
		%	& \leq -2\gamma_1 k \sum_{i=1}^{n}  \bm{\xi}_{i}^2 + 2\gamma_1\beta k \sum_{i=1}^{n}\sum_{j = 1}^{n}{a_{ij}}  \bm{\xi}_{i}^{\text{T}} \bm{\xi}_{j}(t-\tau_{ij})
		%	\end{aligned}
	\begin{aligned} 
		\dot{V}_1 \leq -2\gamma_1 k \sum_{i=1}^{n}  \bm{\xi}_{i}^2 + 2\gamma_1\beta k \sum_{i=1}^{n}\sum_{j = 1}^{n}{a_{ij}}  \bm{\xi}_{i}^{\text{T}} \bm{\xi}_{j}(t-\tau_{ij})
	\end{aligned}
\end{equation}
According to the relationship Eq.(\ref{sum}), it can be obtained
%\begin{equation}
%	\label{Lyapunov_2}
%	\begin{aligned} 
	%		\dot{V}_2 &= \gamma_2\sum_{i=1}^{n}\sum_{j = 1}^{n} a_{ji}   \bm{\xi}_{i}^2 - \gamma_2\sum_{i=1}^{n}\sum_{j = 1}^{n} a_{ji} \bm{\xi}_{i}^2(t-\tau)  \\
	%		& \leq \gamma_2\sum_{i=1}^{n} n  \bm{\xi}_{i}^2 - \gamma_2\sum_{i=1}^{n}\sum_{j = 1}^{n} a_{ji} \bm{\xi}_{i}^2(t-\tau)
	%	\end{aligned}
%\end{equation}

\begin{equation}
	\label{Lyapunov_2}
	\begin{aligned} 
		\dot{V}_2 \leq \gamma_2\sum_{i=1}^{n} n  \bm{\xi}_{i}^2 - \gamma_2\sum_{i=1}^{n}\sum_{j = 1}^{n} a_{ji} \bm{\xi}_{i}^2(t-\tau)
	\end{aligned}
\end{equation}

\begin{equation}
	\label{Lyapunov_3}
	\begin{aligned} 
		\dot{V}_3  \leq \sum_{i=1}^{n} \gamma_3 n\tau \dot{\bm{\xi}}_{i}^2 - \sum_{i=1}^{n}\sum_{j = 1}^{n} a_{ji} \gamma_3    \int_{t-\tau}^{t}{ \dot{\bm{\xi}}_{i}^2(s) }  \text{d}s  
	\end{aligned}
\end{equation}
By \textit{Lemma 2}, it can be obtained
\begin{equation}
	\label{lemma2}
	\begin{aligned} 
		\int_{t-\tau}^{t}{ \dot{\bm{\xi}}_{i}^2(s) }  \text{d}s =  \int_{t-\tau}^{t-\tau_{ji}}{ \dot{\bm{\xi}}_{i}^2(s) }  \text{d}s +  \int_{t-\tau_{ji}}^{t}{ \dot{\bm{\xi}}_{i}^2(s) }  \text{d}s  \\
		\geq \frac{(\bm{\xi}_{i}(t-\tau_{ji}) - \bm{\xi}_{i}(t-\tau))^2}{\tau-\tau_{ji}}     +    \frac{(\bm{\xi}_{i} - \bm{\xi}_{i}(t-\tau_{ji}) )^2}{\tau_{ji}}
	\end{aligned}
\end{equation}
By \textit{Lemma 3}, the lower bound for convex combination of Eq.(\ref{lemma2}) is
\begin{equation}
	\label{lemma3}
	\begin{aligned} 
		\int_{t-\tau}^{t}{ \dot{\bm{\xi}}_{i}^2(s) }  \text{d}s \geq  \frac{1}{2\tau}\bm{\xi}_{i}^2(t-\tau_{ji}) - \frac{1}{2\tau}\bm{\xi}_{i}^2(t-\tau) - \frac{1}{2\tau}\bm{\xi}_{i}^2 
	\end{aligned}
\end{equation}
%\begin{equation}
%	\label{lemma3}
%	\begin{aligned} 
	%		&  \int_{t-\tau}^{t}{ \dot{\bm{\xi}}_{i}^2(s) }  \text{d}s \geq  \frac{(\bm{\xi}_{i}(t-\tau_{ji}) - \bm{\xi}_{i}(t-\tau))^2}{\tau}   \\ 
	%		& \quad + \frac{(\bm{\xi}_{i} - \bm{\xi}_{i}(t-\tau_{ji}))^2}{\tau} \\
	%		& \quad + \frac{(\bm{\xi}_{i}(t-\tau_{ji}) - \bm{\xi}_{i}(t-\tau))^\text{T}(\bm{\xi}_{i} - \bm{\xi}_{i}(t-\tau_{ji}))}{\tau}  \\ 
	%		& = \frac{1}{\tau} \Big[ \bm{\xi}_{i}^2(t-\tau_{ji}) + \bm{\xi}_{i}^2(t-\tau) + \bm{\xi}_{i}^2 - \bm{\xi}_{i}^\text{T}\bm{\xi}_{i}(t-\tau) \\ & \quad -  \bm{\xi}_{i}^\text{T}\bm{\xi}_{i}(t-\tau_{ji}) - \bm{\xi}_{i}^\text{T}(t-\tau_{ji})\bm{\xi}_{i}(t-\tau) \Big] \\
	%		& =  \frac{1}{\tau} \Big[ \frac{1}{2}\bm{\xi}_{i}^2(t-\tau_{ji}) + (\frac{1}{2}\bm{\xi}_{i}(t-\tau_{ji}) - \bm{\xi}_{i}(t-\tau))^2 \\
	%		& \quad + (\frac{1}{2}\bm{\xi}_{i}(t-\tau_{ji})  - \bm{\xi}_{i})^2 + \frac{1}{2}(\bm{\xi}_{i}  - \bm{\xi}_{i}(t-\tau))^2 \\
	%		& \quad  - \frac{1}{2}\bm{\xi}_{i}^2(t-\tau) - \frac{1}{2}\bm{\xi}_{i}^2    \Big]\\
	%		& \geq  \frac{1}{2\tau}\bm{\xi}_{i}^2(t-\tau_{ji}) - \frac{1}{2\tau}\bm{\xi}_{i}^2(t-\tau) - \frac{1}{2\tau}\bm{\xi}_{i}^2 
	%	\end{aligned}
%\end{equation}
By using \textit{Lemma 1} twice and Eq.(\ref{sum}), it can be obtained
\begin{equation}
	\label{lemma1}
	\begin{aligned} 
		\dot{\bm{\xi}}_{i}^2 \leq 2 n\sum_{j = 1}^{n}{a_{ij}}k^2\bm{\xi}_{i}^2 + 2\beta^2 n \sum_{j = 1}^{n}a_{ij}k^2\bm{\xi}_{j}^2(t-\tau_{ij})
	\end{aligned}
\end{equation}
Substituting Eqs.(\ref{lemma3})(\ref{lemma1}) into Eq.(\ref{Lyapunov_3}) and using Eq.(\ref{sum}) obtains
%\begin{equation}
%	\label{Lyapunov_3_scaling}
%	\begin{aligned} 
	%		\dot{V}_3  &\leq \sum_{i=1}^{n} \gamma_3 n\tau \dot{\bm{\xi}}_{i}^2 - \sum_{i=1}^{n}\sum_{j = 1}^{n} a_{ji} \gamma_3    \int_{t-\tau}^{t}{ \dot{\bm{\xi}}_{i}^2(s) }  \text{d}s    \\
	%		&\leq 2 \gamma_3 \tau n^2 \sum_{i=1}^{n}\sum_{j = 1}^{n}{a_{ij}} \kappa^2 \bm{\xi}_{i}^2  + \sum_{i=1}^{n}\sum_{j = 1}^{n} 	a_{ji}\frac{\gamma_3}{2\tau}\bm{\xi}_{i}^2(t-\tau) \\
	%		&\quad + 2 \gamma_3 \tau n^2 \beta^2 \sum_{i=1}^{n}\sum_{j = 1}^{n}{a_{ij}}  \kappa^2 \bm{\xi}_{j}^2(t-\tau_{ij}) \\
	%		& \quad - \sum_{i=1}^{n}\sum_{j = 1}^{n} a_{ij}\frac{\gamma_3}{2\tau}\bm{\xi}_{j}^2(t-\tau_{ij}) + \sum_{i=1}^{n}\frac{\gamma_3}{2\tau}n\bm{\xi}_{i}^2   \\
	%		&\leq \sum_{i=1}^{n} ( 2 \gamma_3 \tau n^3 \kappa^2 + \frac{\gamma_3}{2\tau}n I_6) \bm{\xi}_{i}^2 + \sum_{i=1}^{n}\sum_{j = 1}^{n} a_{ji}\frac{\gamma_3}{2\tau}\bm{\xi}_{i}^2(t-\tau) \\
	%		&\quad + \sum_{i=1}^{n}\sum_{j = 1}^{n}{a_{ij}} ( 2 \gamma_3 \tau n^2 \beta^2 \kappa^2 - \frac{\gamma_3}{2\tau}I_6) \bm{\xi}_{j}^2(t-\tau_{ij}) \\
	%	\end{aligned}
%\end{equation}
\begin{equation}
	\label{Lyapunov_3_scaling}
	\begin{aligned} 
		\dot{V}_3  & \leq \sum_{i=1}^{n}\sum_{j = 1}^{n}{a_{ij}} ( 2 \gamma_3 \tau n^2 \beta^2 k^2 - \frac{\gamma_3}{2\tau}I_6) \bm{\xi}_{j}^2(t-\tau_{ij})  \\
		& + \sum_{i=1}^{n} ( 2 \gamma_3 \tau n^3 k^2 + \frac{\gamma_3}{2\tau}n I_6) \bm{\xi}_{i}^2 \\
		& + \sum_{i=1}^{n}\sum_{j = 1}^{n} a_{ji}\frac{\gamma_3}{2\tau}\bm{\xi}_{i}^2(t-\tau)
	\end{aligned}
\end{equation}
Integrating Eqs.(\ref{Lyapunov_1})(\ref{Lyapunov_2})(\ref{Lyapunov_3_scaling}) obtains a negative definite quadratic form
\begin{equation}
	\label{dotLyapunov}
	\begin{aligned} 
		\dot{V}  &\leq - \alpha_1 \sum_{i=1}^{n}  \bm{\xi}_{i}^2  - \alpha_2 \sum_{i=1}^{n}\sum_{j = 1}^{n} a_{ij} \bm{\xi}_{j}^2(t-\tau_{ij}) \\
		& \quad - \alpha_3 \sum_{i=1}^{n}\sum_{j = 1}^{n} a_{ji} \bm{\xi}_{i}^2(t-\tau) \\
	\end{aligned}
\end{equation}
\begin{equation}
	\label{conditions}
	\begin{aligned} 
		\begin{cases}  \alpha_1 = 2\gamma_1k - 2 \gamma_3 \tau n^3k^2 - (\gamma_2  + \frac{\gamma_3}{2\tau} + \gamma_1\beta k )n  I_6 \\\\ 
			\alpha_2 = (\frac{\gamma_3}{2\tau} - \gamma_1\beta k)I_6- 2 \gamma_3 \tau n^2 \beta^2 k^2   \\\\
			\alpha_3 = (\gamma_2 -  \frac{\gamma_3}{2\tau})I_6 \end{cases}
	\end{aligned}
\end{equation}
In order to ensure $\dot{V} \leq 0$, $\alpha_1$, $\alpha_2$, and $\alpha_3$ must be semi-positive definite matrices, which denotes $\alpha_1, \alpha_2, \alpha_3 > 0$ since they are diagonal matrices.
For a system, the number of robots $n$ and communication delay bound $\tau$ are determined.
The control law is designed by finding the coefficients $k,\beta > 0$ such that there exists $\gamma_1,\gamma_2,\gamma_3 > 0$ that satisfies $\alpha_1, \alpha_2, \alpha_3 > 0$.
A simple design criterion is that the coefficients $k,\beta$ are related to the communication delay bound.
When the bound is large, the control coefficients should be smaller.
Once the appropriate coefficients are found, it can be concluded that $V \geq 0$ and $\dot{V} \leq 0$.
$\dot{V} = 0$ if and only if $\bm{\xi}_{i} = 0$ for $i = 1, \dots, n$.

By using control law Eq.(\ref{control_law}), the joint position errors will converge to zero, which in turn drives the end-effector pose errors $\bm{e}_i^{\text{d}} -  \bm{e}_i$ to zero.
Furthermore, the interaction wrenches can approach the desired values according to Eq.(\ref{joint_space}). 
Therefore, the unnecessary interaction wrenches can be reduced.

\textit{Remark 1}: The control law does not correct the deviation of the planned trajectory.
For example, if each robot has a pose deviation or velocity deviation, but the deviations are the same, then the input by Eq.(\ref{control_law}) is zero since $\bm{h}_i = \bm{h}_i^{\text{ref}}$ and $\bm{h}_j = \bm{h}_j^{\text{ref}}$.
This is normal since only local information is used and thus cannot correct global errors.

\subsection{Simulations}

\begin{figure}[t]
	\includegraphics[width=\columnwidth]{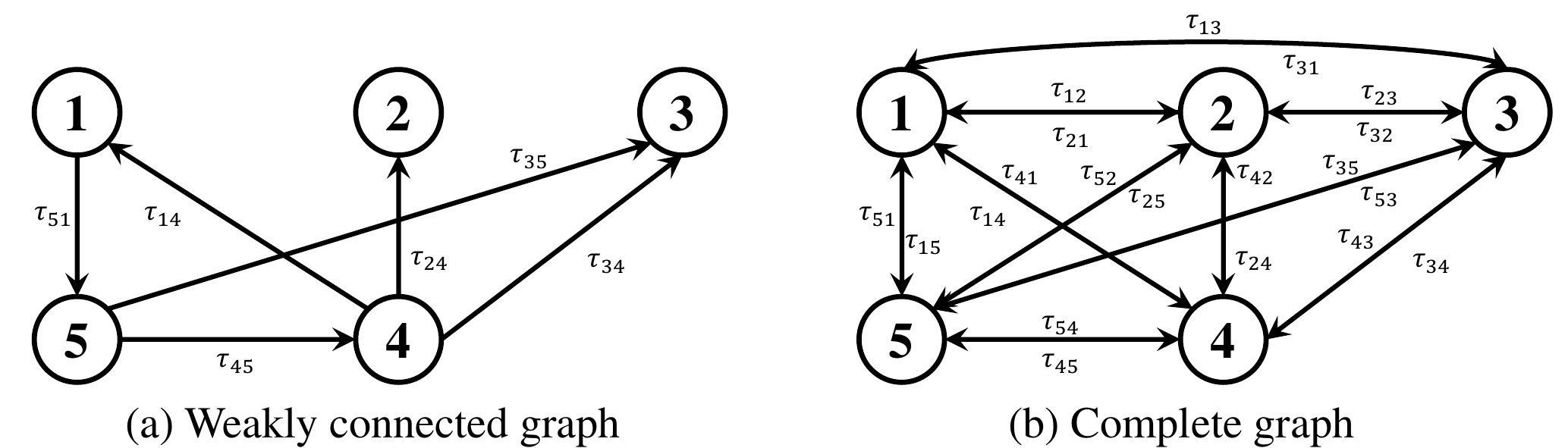}
	\centering
	\caption{
		The communication topology for multiple robots. (a) Weakly connected graph. (b) Complete graph.
	}
	\label{topo}       
\end{figure}

\begin{figure}[t]
	\includegraphics[width=\columnwidth]{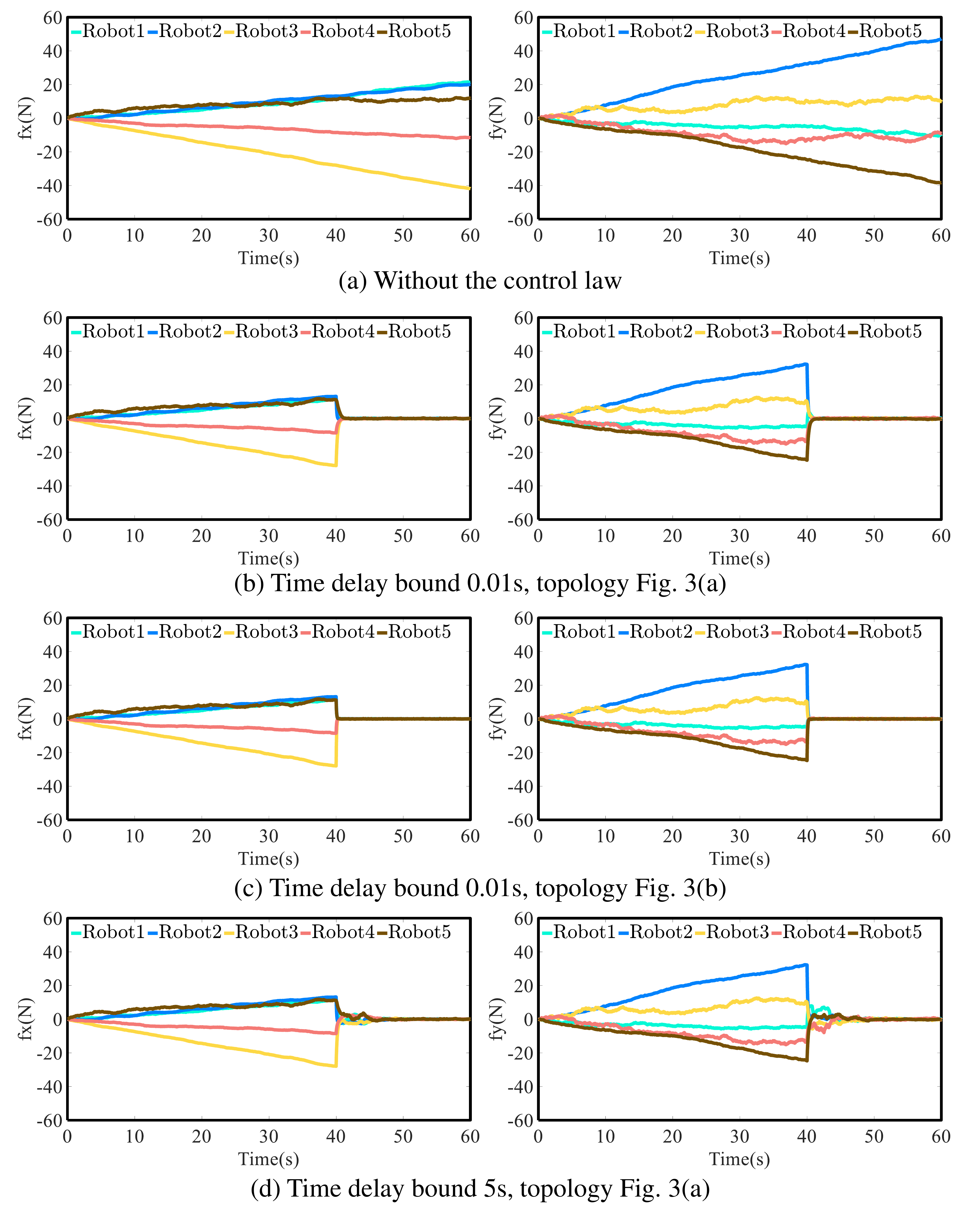}
	\centering
	\caption{
		The interaction wrenches between robots and the object. (a) Without the proposed control law. (b)-(d) With the proposed control law.
	}
	\label{simulation}       
\end{figure}

\begin{figure}[t]
	\includegraphics[width=\columnwidth]{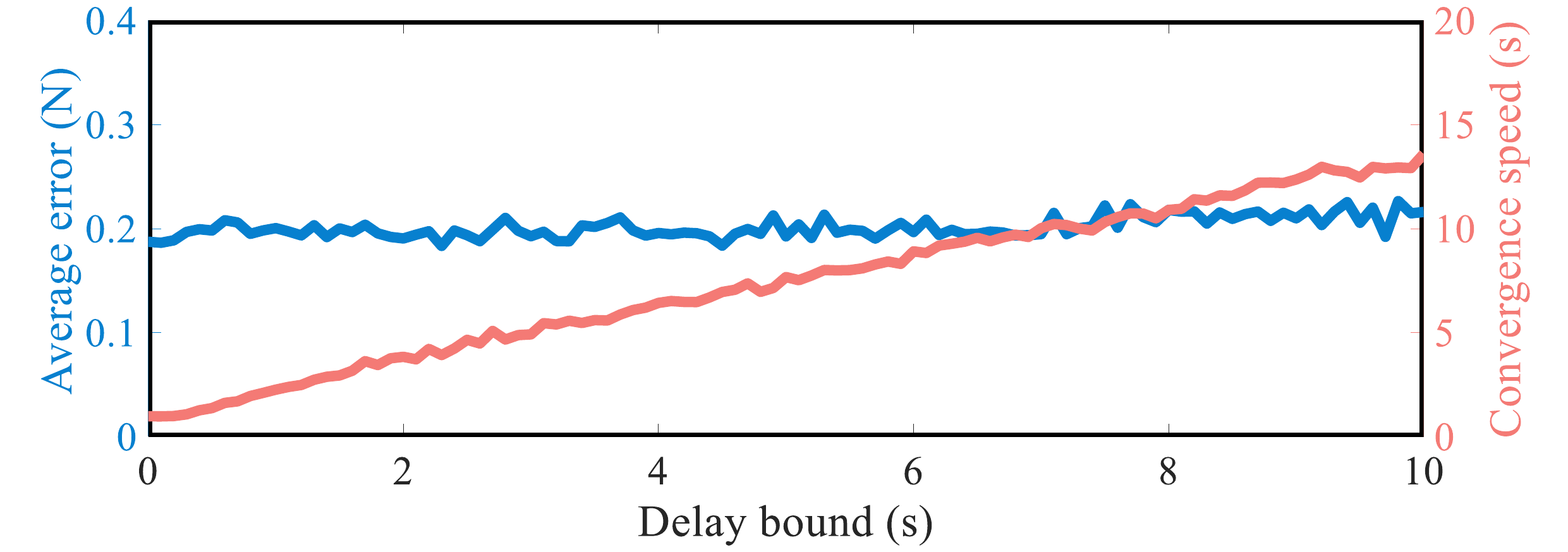}
	\centering
	\caption{
		The effect of the delay upper bound on the control law.
	}
	\label{delayvarying}       
\end{figure}

\begin{figure}[t]
	\includegraphics[width=\columnwidth]{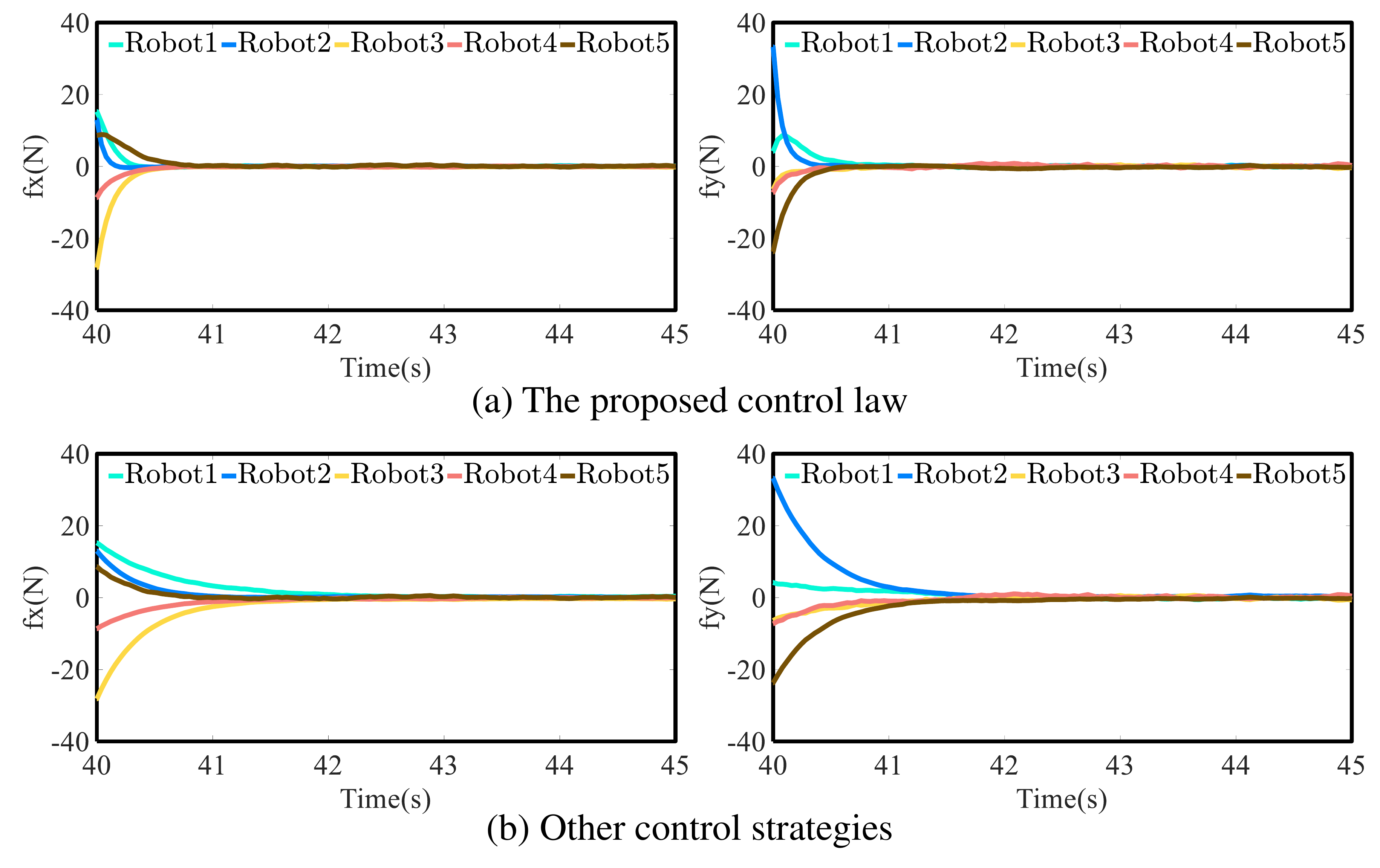}
	\centering
	\caption{
		A comparison of the convergence speed with other methods.
	}
	\label{cmp1}       
\end{figure}

\begin{figure}[t]
	\includegraphics[width=\columnwidth]{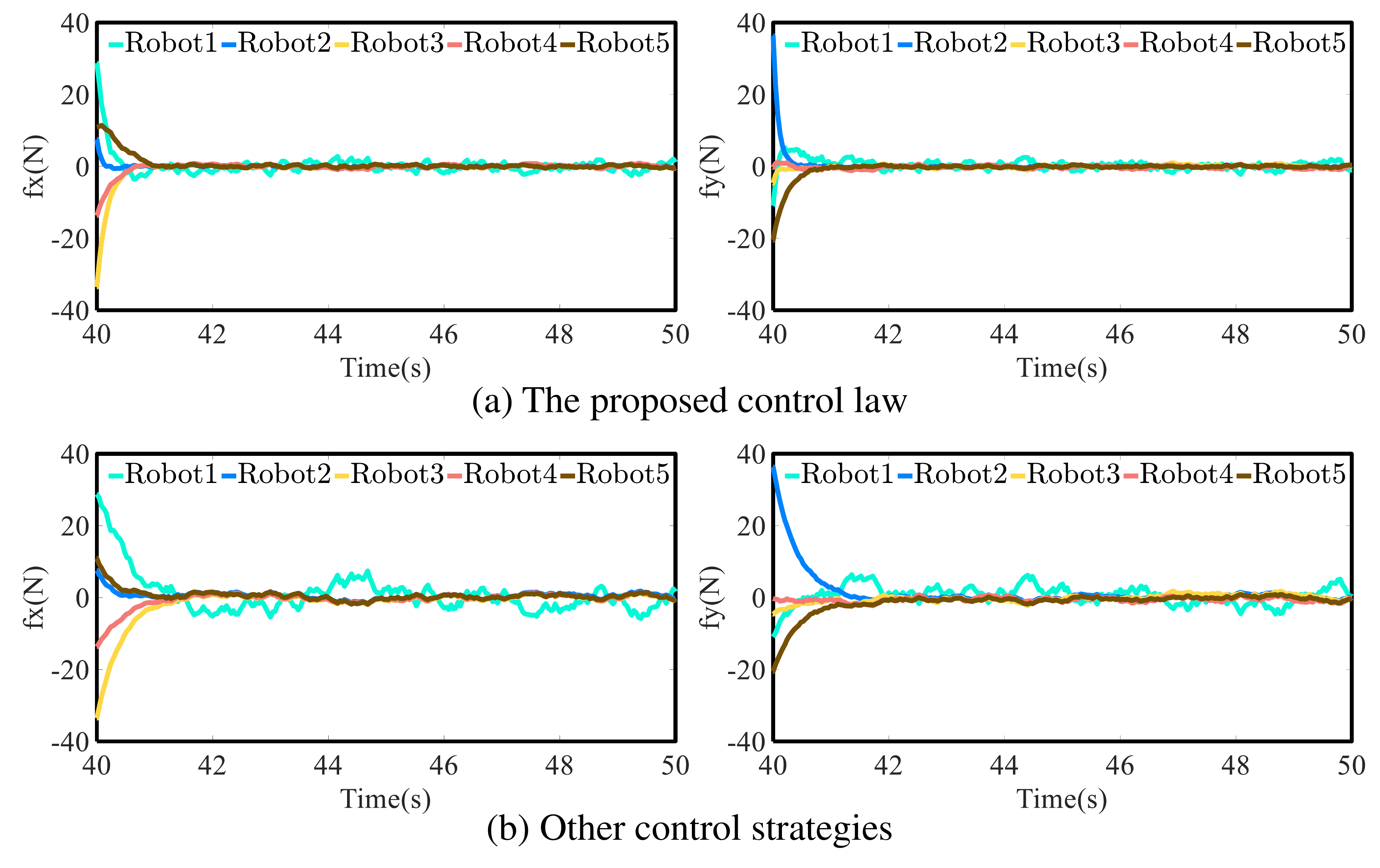}
	\centering
	\caption{
		A comparison of the robustness with other methods.
	}
	\label{cmp2}       
\end{figure}

\begin{figure}[t]
	\includegraphics[width=0.95\columnwidth]{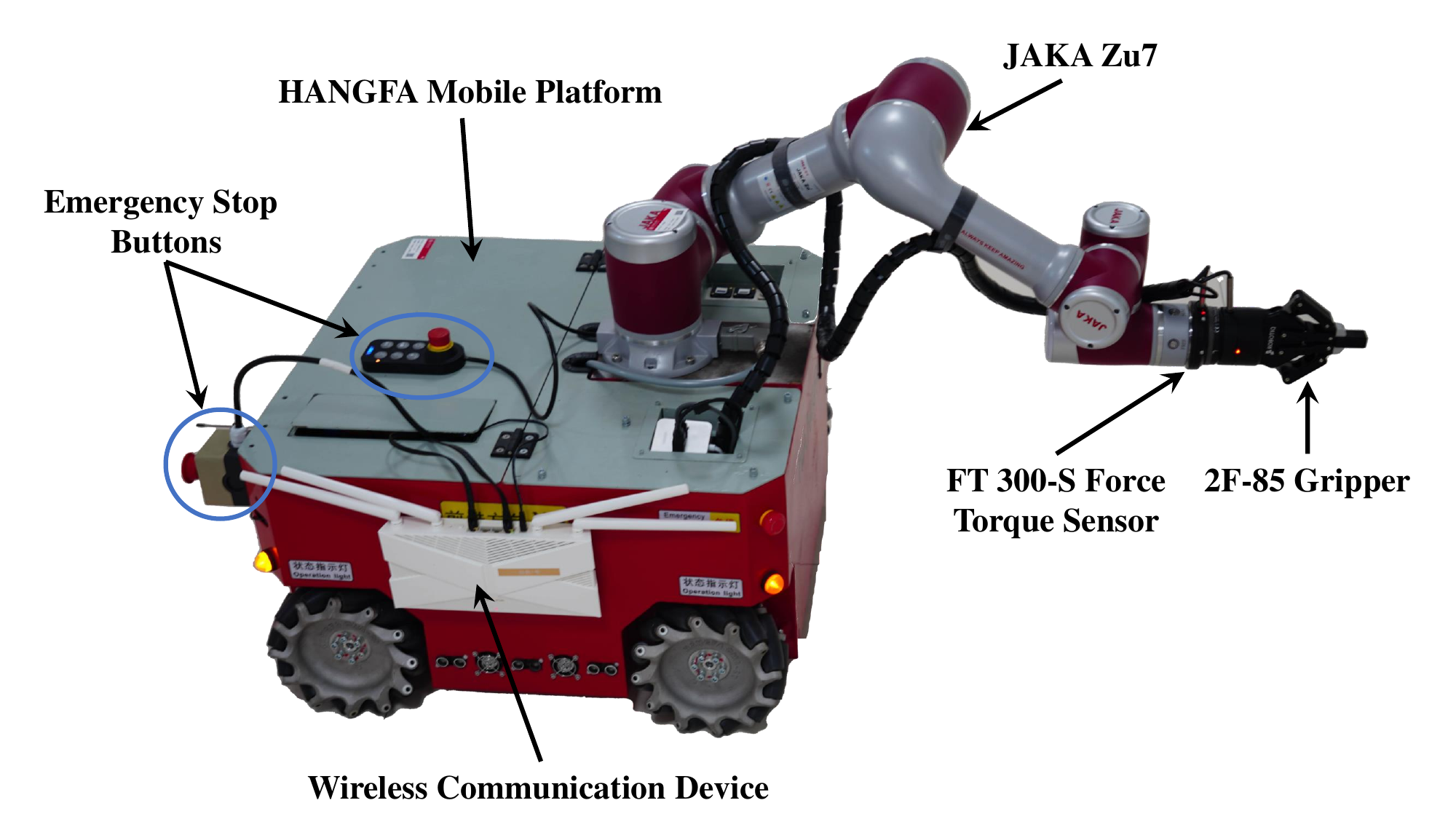}
	\centering
	\caption{
		The physical model of a mobile manipulator robot used in experiments.
	}
	\label{physical_system}       
\end{figure}

\begin{figure*}[t]
	\includegraphics[width=0.95\textwidth]{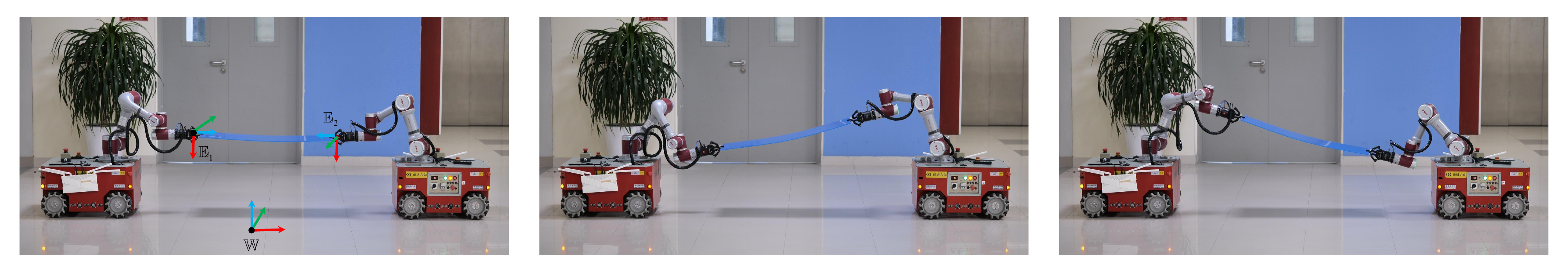}
	\centering
	\caption{ Two mobile manipulator robots cooperatively manipulate an object to perform a reciprocating rotational motion, with the object's angle ranging between -15 degrees and 15 degrees. }
	\label{exp_rotate}       
\end{figure*}

\begin{figure*}[t]
	\includegraphics[width=0.75\textwidth]{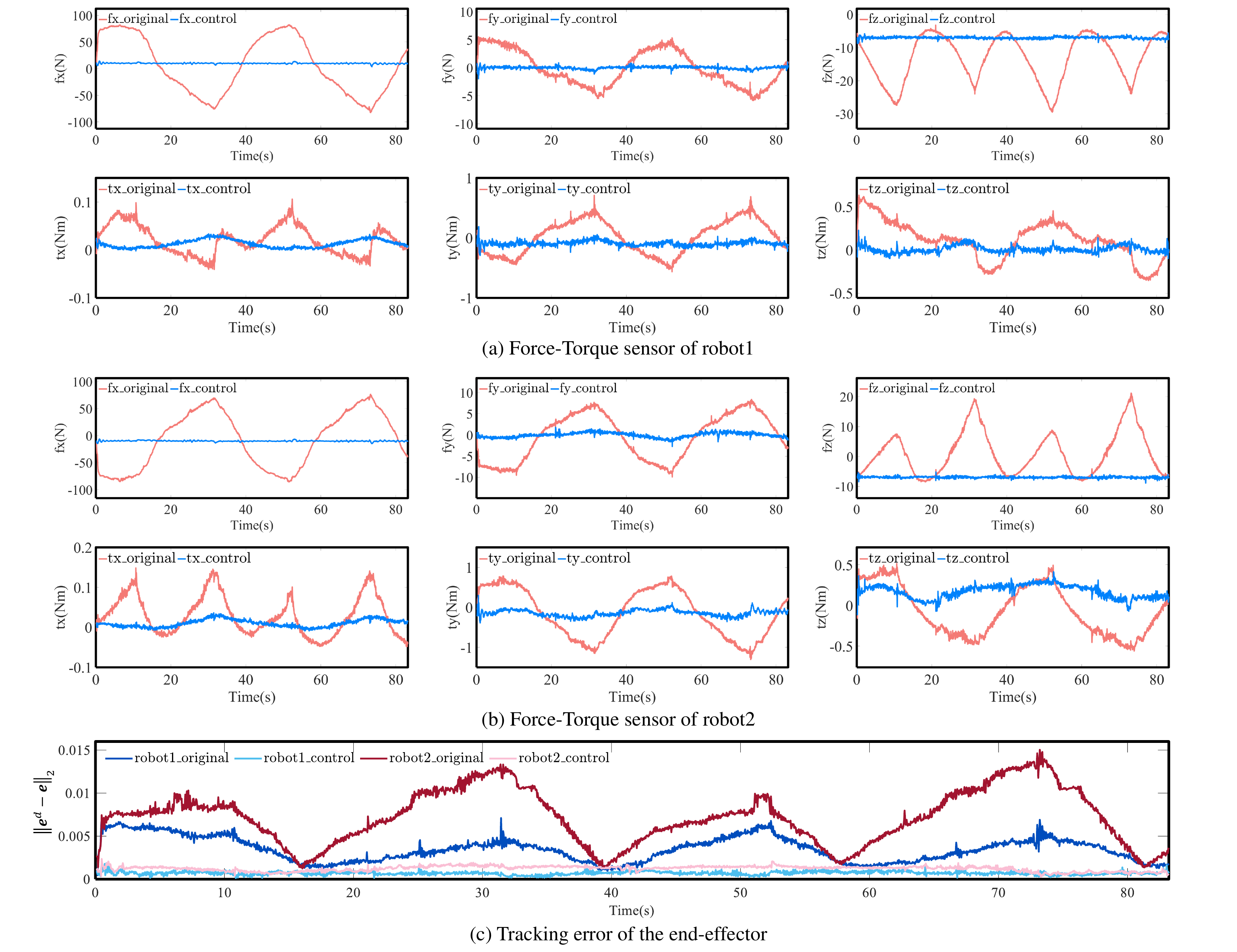}
	\centering
	\caption
	{ Comparison between applying and not applying the control law.
		The interaction wrenches in the world frame without (with) applying the control law is represented by the red (blue) curve.	
	}
	\label{exp_rotate_data}       
\end{figure*}

\begin{table}[t]
	\centering\small
	\begin{threeparttable}
		\caption{\label{speeds}The actual end velocity (m/s) of each robot}
		\begin{tabular}{m{0.07\textwidth}<{\centering}m{0.35\textwidth}<{\centering}}
			\toprule
			$\dot{\bm{e}}_1$ & [$0.1+0.05\Omega(t)$, $0.1+0.1\Omega(t)$]\tnote{T}  \\
			$\dot{\bm{e}}_2$ & [$0.1+0.01sin(t)$, $0.08$]\tnote{T}  \\
			$\dot{\bm{e}}_3$ & [$0.12$, $0.1+0.2\Omega(t)$]\tnote{T}  \\
			$\dot{\bm{e}}_4$ & [$0.11+0.05\Omega(t)sin(t)$, $0.1+0.2\Omega(t)$]\tnote{T}  \\
			$\dot{\bm{e}}_5$ & [$0.1+0.1\Omega(t)$, $0.11$]\tnote{T}  \\
			\bottomrule 
		\end{tabular} \small
	\end{threeparttable}
\end{table}

\begin{figure*}[t]
	\includegraphics[width=0.95\textwidth]{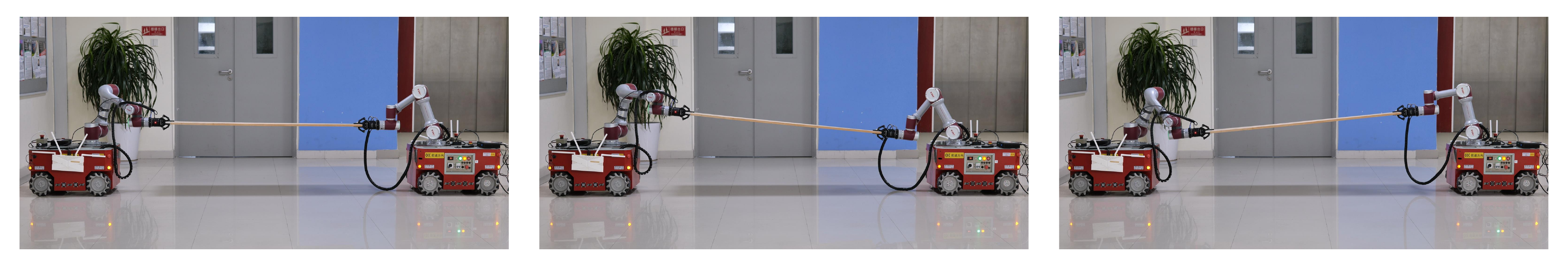}
	\centering
	\caption{ Two mobile manipulator robots cooperatively manipulate an object to perform a reciprocating rotational motion.}
	\label{exp_rotate_heavier}       
\end{figure*}

\begin{figure*}[t]
	\includegraphics[width=0.8\textwidth]{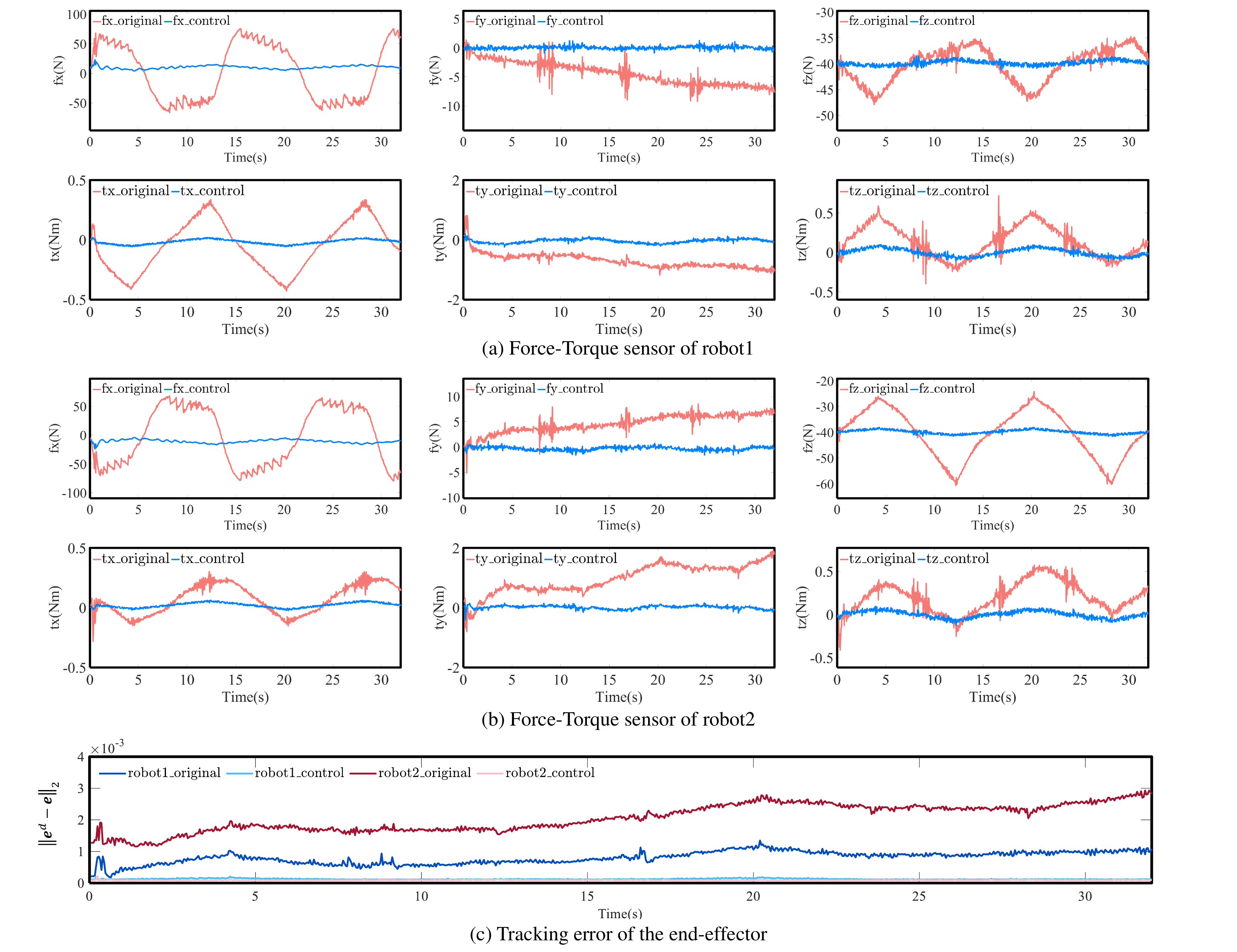}
	\centering
	\caption
	{ Comparison between applying and not applying the control law.
		The interaction wrenches in the world frame without (with) applying the control law is represented by the red (blue) curve.
	}
	\label{exp_rotate_heavier_data}       
\end{figure*}

\begin{figure*}[t]
	\includegraphics[width=0.95\textwidth]{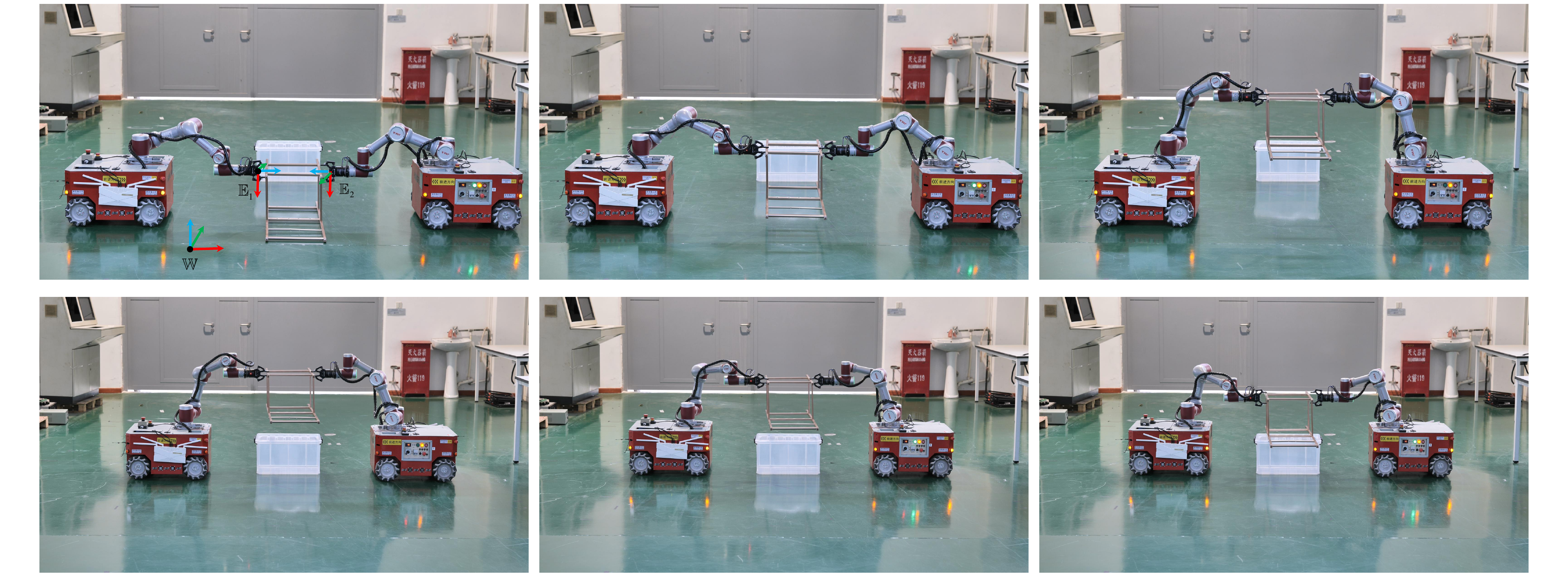}
	\centering
	\caption{ Two mobile manipulator robots cooperatively transport an object.}
	\label{exp_trans}       
\end{figure*}

\begin{figure*}[t]
	\includegraphics[width=0.8\textwidth]{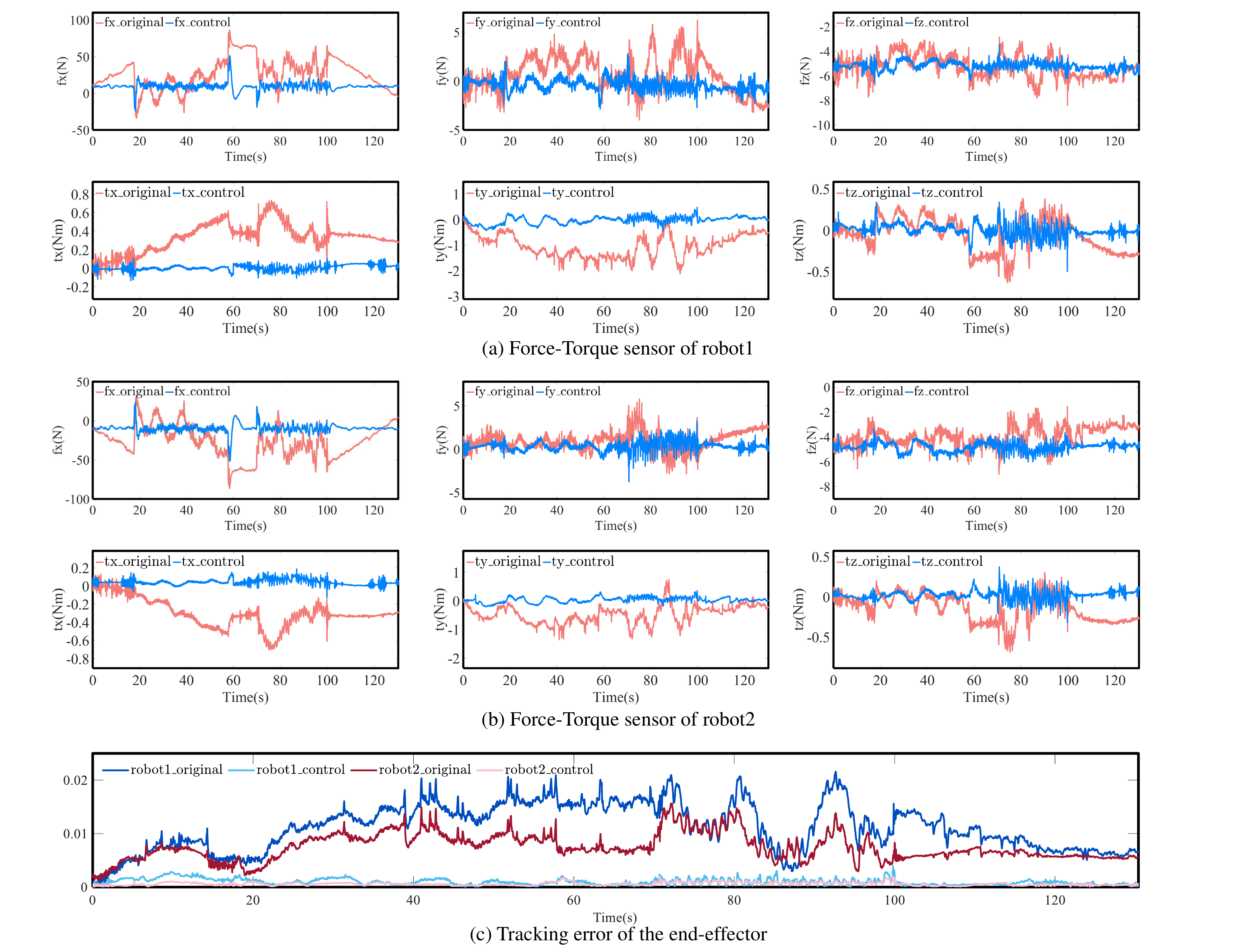}
	\centering
	\caption{ Comparison between applying and not applying the control law. The force-torque sensor data without (with) applying the control law is represented by the red (blue) curve.}	
	\label{exp_trans_data}       
\end{figure*}

In order to verify the stability of the control law, numerical simulations are performed.
Consider a system of 5 robots manipulating objects on a 2D plane.
Assume that the ends of all robots move on the same 2D plane, and $\dot{\bm{e}}_{i}^{\text{ref}} = [0.1,0.1]^\text{T}$m/s, $\bm{h}_i^{\text{ref}} = [0, 0]^\text{T}$, $i = 1,\dots, 5$.
The whole process is 60 seconds.
Let $\Omega(\cdot)$ denote a random number between -1 and 1.
The stiffness of the object is assumed to be $K = \text{diag}\left\{10.5, 9.5\right\}$.

Due to some uncertainties and the deviation between the planned robot model and the actual robot model, it is impossible to achieve the planned velocity.
The actual end velocities of robots are shown in Table \ref{speeds}.
If the deviation is not corrected by employing the control law, it will result in increased interaction wrenches, as illustrated in Fig.~\ref{simulation}(a).
In the simulation, two types of errors are introduced.
One type is random velocity error, and the other is velocity bias.
The bias can be caused by the relative pose errors of the robots.
Both types of errors are very common in practice, and may cause increased interaction wrenches.

The proposed control law Eq.(\ref{control_law}) is verified by enrolling it after 40s in the process described above.
The communication topology is shown in Fig.~\ref{topo}.
As described before, time delays are non-uniform, asymmetric, and time-varying, and they are assumed to be $\tau_{12}=||0.01\text{sin}(t)||, \tau_{13}=||0.01\Omega(t)||, \tau_{14}=||0.01\text{cos}(t)||, \tau_{15}=||0.01\Omega(t)||, \tau_{21}=\frac{0.02}{t}, \tau_{23}=\frac{0.02}{t^2}, \tau_{24}=\text{e}^{-t}, \tau_{25}=t\text{e}^{-t}, \tau_{31} = ||0.01\Omega(t)||, \tau_{32}=||0.01\Omega(t)||, \tau_{34}= 0.002\text{ln}t, \tau_{35}= \frac{0.02}{\text{ln}t}, \tau_{41}=\frac{0.02}{\text{ln}t}, \tau_{42}=||0.01\text{sin}(t)||, \tau_{43}=||0.01\text{cos}(t)||, \tau_{45}=||0.01\Omega(t)||, \tau_{51}=||0.01\Omega(t)||, \tau_{52}=||0.01\Omega(t)||, \tau_{53}=||0.01\Omega(t)||, \tau_{54}=||0.01\Omega(t)||$.
The control coefficients are set as $k = 0.5, \beta = 0.1$.
The control frequency of the robots is 25 Hz.
According to these time delays and $t\in [40,60]$, the bound $\tau$ is 0.01s.
These parameters substituted into Eq.(\ref{conditions}) satisfy $\alpha_1, \alpha_2, \alpha_3 > 0$.
Therefore, the control law Eq.(\ref{control_law})  is stable.
The result for topology Fig.~\ref{topo}(a) is illustrated in Fig.~\ref{simulation}(b).
After 40s, the control law is enrolled, and the wrench is quickly adjusted to the desired value.
Although we use five robots as an example, the control law can easily be extended to more robots.
The computation of the control law Eq.(\ref{control_law}) itself is not complex.

To test the effect of communication graphs with different connectivity on the control law, the graph in Fig.~\ref{topo}(b) is used.
The parameters and other conditions are identical, and the result is shown in Fig.~\ref{simulation}(c).
If stability is defined as the errors being less than 0.5N, Fig.~\ref{simulation}(b) requires 1.12s after applying the control law, while Fig.~\ref{simulation}(c) requires 0.32s.
Since Fig.~\ref{topo}(b) is a complete graph with stronger connectivity than Fig.~\ref{topo}(a), the robots can communicate and share information more directly and efficiently.
Therefore, the convergence speed is much faster.
For the proposed distributed control law, when the parameters are the same, stronger graph connectivity leads to faster convergence.

To test the effect of communication delay on the control law, the upper bound of the delay is adjusted to 5s.
Other parameters and conditions are the same with those in Fig.~\ref{simulation}(b).
The result is shown in Fig.~\ref{simulation}(d).
After the control law is applied, it takes 7.84s to reach stability.
To further investigate the impact of time delays, tests are conducted with various delay upper bounds.
Since both delays and disturbances are random errors, each test is repeated 100 times, and the average values for convergence speed and steady-state errors are recorded.
The results are shown in Fig.~\ref{delayvarying}.
For different delays even large delays, the control law can stabilize in the end.
It can be seen that under the same parameters and conditions, a larger delay upper bound leads to slower convergence speed, but has little effect on the steady-state errors.
The proposed control law is robust to communication delays.
At a steady state, the control law can reduce the errors to a low level.
However, as the communication delay increases, the information received by each robot becomes more delayed, which can affect the dynamic performance.

Due to the difficulty in fully replicating existing approaches, the basic leader-follower control strategies similar to \cite{wu2016collaboration, ren2024hybrid} were conducted.
The setup is consistent with Fig.~\ref{simulation}.
It is assumed that the time delay is zero.
The first comparison is the convergence speed.
The parameters in the control law are set as the same and the result (40s to 45s) is shown in Fig.~\ref{cmp1}.
When the proposed control law is applied at 40s, it takes only 0.84s to reach stability.
Applying other control strategies takes 2.52s to reach stability.
The convergence speed improves by 66.67\%.
The second comparison is robustness.
A large velocity disturbance is applied to one robot to test the impact on the control laws.
Let the end-effector velocity of Robot 1 become [$0.1+\Omega(t)$, $0.1+\Omega(t)$]$^\text{T}$m/s.
The result (40s to 50s) is shown in Fig.~\ref{cmp2}.
The average error of the proposed control law from 45s to 50s is 1.2724N, while the average error of other control strategies is 2.9547N.
When the disturbance is large, the error is reduced by 56.94\%.
The comparison shows that the proposed control law has significant advantages in both convergence speed and robustness.

\section{Experiments and Discussions}

\subsection{Experimental Setup}

The proposed control law is verified in physical experiments.
Each robot is composed of a HANGFA omnidirectional wheeled platform and a JAKA Zu7 manipulator.
Each robot is equipped with a Robotiq FT 300-S Force Torque Sensor and a Robotiq 2F-85 Gripper as the end-effector, and an onboard Next Unit of Computing (NUC) computer with Intel Core i7-1165G7 CPU at 2.80 GHz (4 cores) and 16GB RAM.
Each onboard computer independently calculates the motion of its respective robot according to the control law.
The control frequency of the robots is 25 Hz.
Besides, all robots are in the same local network for communication.
The communication framework is implemented and achieved through ROS-melodic.
Generally, the communication delays are less than 0.01s.
The physical robot is shown in Fig.~\ref{physical_system}.

\subsection{Experimental Results}

The control law is validated through experiments involving two robots cooperating to manipulate objects.
In the first experiment, the robots collaborate to perform reciprocating rotational motions with an acrylic sheet, and the oscillation angle ranges from -15 to 15 degrees.
The sheet measures $1m$ in length, $0.4m$ in width, and $0.003m$ in thickness, with a weight of $14N$.
The expected interaction wrenches of robots in the world frame are $\bm{h}_{1}^{\text{ref}} = [\bm{f}_1^{\text{ref}},\bm{t}_1^{\text{ref}}]^{\text{T}}$, $\bm{f}_1^{\text{ref}} = [10,0,-7]^{\text{T}}N$, $\bm{t}_1^{\text{ref}} = [0,0,0]^{\text{T}}Nm$, $\bm{f}_2^{\text{ref}} = [-10,0,-7]^{\text{T}}N$, $\bm{t}_2^{\text{ref}} = [0,0,0]^{\text{T}}Nm$.
The weight is evenly distributed in the $z$-direction by two robots, and a $10N$ force in the $x$-direction is applied to maintain the stability of the object.
The coefficients in the control law Eq.(\ref{control_law}) are set as $K_0 =\text{diag}\left\{ 5\times10^5, 2\times10^5, 2\times10^5, 10^2, 10^2, 10^2\right\}$ and $k = 0.5, \beta = 0.2$.

The process of the experiment is illustrated in Fig.~\ref{exp_rotate}, and the motion duration is 83.2s.
The experiment is repeated twice, once without the control law and once with.
The results of interaction wrenches are shown in Fig.~\ref{exp_rotate_data}.
Without the control law, the interaction wrenches could not be maintained at the desired values.
Especially in the $x$-direction, the forces are very large due to the pushing and pulling between the robots when cooperatively manipulating the object.
Additionally, in the $z$-direction, the robots could not evenly share the load.
In contrast, with the control law applied, the interaction wrenches are maintained at the desired values in all six directions.
Compared to the scenario without the control law, implementing the control law reduces the maximum errors by [93.37, 66.62, 92.38, 68.75, 58.82, 64.24]\% and the average errors by [98.68, 93.49, 96.58, 55.36, 61.96, 84.89]\% for robot1 in the corresponding directions, and the maximum errors by [94.38, 72.55, 93.30, 75.74, 70.18, 26.05]\% and the average errors by [98.56, 90.15, 97.30, 74.94, 73.36, 37.10]\% for robot2 in the corresponding directions.
The tracking errors of the end-effector $||\bm{e}^{\text{d}} -  \bm{e}||_2$ are shown in Fig.~\ref{exp_rotate_data}(c).
Applying the control law reduces the average tracking errors for robot 1 and robot 2 by 79.62\% and 82.93\%, respectively.
As previously mentioned, due to the tight contact between the robot and the object, even small errors can lead to significant interaction wrenches.

Further, we conduct experiments with a more rigid object, as shown in Fig.~\ref{exp_rotate_heavier}.
The object is replaced with a wooden board weighing 80N.
The corresponding results are shown in Fig.~\ref{exp_rotate_heavier_data}.
It can be seen that the proposed control law is still effective in reducing interaction wrenches and maintaining them at the desired values.
Applying the control law reduces the average tracking errors for robot 1 and robot 2 by 92.52\% and 91.82\%, respectively.

In the second experiment, the robots cooperatively transport an object to a desired location.
The object is an aluminum frame, weighing $10N$.
The entire motion of robots is planned in advance, with the expected interaction wrenches $\bm{f}_1^{\text{ref}} = [10,0,-5]^{\text{T}}N$, $\bm{t}_1^{\text{ref}} = [0,0,0]^{\text{T}}Nm$, $\bm{f}_2^{\text{ref}} = [-10,0,-5]^{\text{T}}N$, $\bm{t}_2^{\text{ref}} = [0,0,0]^{\text{T}}Nm$.
The coefficients in the control law Eq.(\ref{control_law}) are set as $K_0 =\text{diag}\left\{ 10^6, 10^5, 10^5, 10^2, 10^2, 10^2\right\}$ and $k = 0.5, \beta = 0.2$.

The process of the experiment is illustrated in Fig.~\ref{exp_trans}, and the motion duration is 130.4s.
From 17s to 57s, the platforms move forward simultaneously to avoid exceeding the reachable space of the manipulators.
From 70s to 100s, the platforms move laterally simultaneously to transport the object.
As shown in Fig.~\ref{exp_trans_data}, it can be seen that without the control law, the interaction wrenches vary greatly, especially in the $f_x$ directions.
Compared to the scenario without the control law, implementing the control law reduces the maximum errors by [45.97, 54.18, 53.56, 81.85, 77.39, 21.52]\% and the average errors by [85.92, 62.72, 60.49, 93.13, 88.80, 67.99]\% for robot1 in the corresponding directions, and the maximum errors by [45.47, 34.97, 52.53, 73.38, 74.41, 46.05]\% and the average errors by [86.38, 58.44, 69.38, 85.51, 83.32, 72.44]\% for robot2 in the corresponding directions.
The tracking errors of the end-effector are shown in Fig.~\ref{exp_rotate_data}(c).
Applying the control law reduces the average tracking errors for robot 1 and robot 2 by 91.39\% and 92.15\%, respectively.

When only the manipulators are moving, the interaction wrenches are well maintained.
However, when the mobile platforms are moving (from 17s to 57s and from 70s to 100s), the force oscillation range becomes significantly larger.
Additionally, some impacts occur due to the starting and stopping of the mobile platform.
From this, it is evident that the accuracy of the manipulator is much higher than that of the omnidirectional wheel.
Therefore, the control law is designed utilizing the motion of manipulators to correct errors.

\subsection{Discussion and Limitation}

The stability of the proposed control law has been proven, and subsequent simulations and experiments have demonstrated that the proposed control law effectively maintains the desired interaction wrenches.
Due to various factors in physical robots, such as kinematic parameter uncertainties and external disturbances, deviations may occur between the actual and planned results.
This can lead to excessive interaction wrenches in cooperative object manipulation.
The proposed control law can be used to address these situations.
Compared to other control methods, the proposed control law only utilizes local sensor information, and does not require accurate dynamic parameters or torque control.
These simple conditions make the control law applicable in most scenarios and more suitable for real-world systems.
Additionally, the communication delay between robots is considered, and it has been proven that the control law is robust against delays.

However, the proposed control strategy also has some limitations.
First, the stiffness of the object is required.
This parameter often requires empirical values or preliminary experimental determination.
Improper setting of control coefficients may lead to instability or divergence of the control law.
Second, since only the joint velocity is considered, the control law may fail in scenarios where the trajectory is demanding, such as when the velocity of the object changes frequently and largely.

\section{Conclusion and Future Work}

This paper aims to reduce the interaction wrenches during the cooperative object manipulation of multiple mobile manipulator robots.
A distributed motion control law based on local information is proposed.
The stability of the control law has been rigorously proven.
The effects of graph connectivity and communication delays on the control law have been studied in simulations.
The effectiveness of the control law has been validated in physical experiments.
In conclusion, the proposed control law is easily implemented in physical systems and general to different robots.

Regarding future work, on the one hand, we will further refine the control strategy.
The dynamics model of the robot will be considered to reduce impact forces when the robot's velocity changes.

On the other hand, we will study the rigorous proof of the control law's robustness to disturbances and parameter uncertainties.

\bibliographystyle{elsarticle-num-names}
\bibliography{References}

\end{document}